\documentclass{tADR2e}

\begin{document}

\jvol{36} \jnum{05-06} \jyear{2022} \jmonth{February}


\title{A survey of multimodal deep generative models}

\author{Masahiro Suzuki$^{a}$$^{\ast}$\thanks{$^\ast$ Email: masa@weblab.t.u-tokyo.ac.jp
\vspace{6pt}} and Yutaka Matsuo$^{a}$\\\vspace{6pt}  $^{a}${{\it Department of Technology Management for Innovation, The University of Tokyo, Tokyo, Japan.}}}

\maketitle
\begin{abstract}
Multimodal learning is a framework for building models that make predictions based on different types of modalities. 
Important challenges in multimodal learning are the inference of shared representations from arbitrary modalities and cross-modal generation via these representations; however, achieving this requires taking the heterogeneous nature of multimodal data into account.

In recent years, deep generative models, i.e., generative models in which distributions are parameterized by deep neural networks, have attracted much attention, especially variational autoencoders, which are suitable for accomplishing the above challenges because they can consider heterogeneity and infer good representations of data. Therefore, various multimodal generative models based on variational autoencoders, called multimodal deep generative models, have been proposed in recent years.
In this paper, we provide a categorized survey of studies on multimodal deep generative models.
\end{abstract}

\begin{keywords}
deep generative models, multimodal learning
\end{keywords}\medskip

\section{Introduction}
We perceive various kinds of sensory information from our external world, such as vision, sounds, and smells. These different types of information are called different {\it modalities}, and it is known that we can develop a more reliable understanding of the world through multiple modalities, or {\it multimodal} information~\cite{stein1993merging}.
In recent years, multimodal learning \cite{baltruvsaitis2018multimodal} has been studied in the field of artificial intelligence and machine learning, which aims to build models that make predictions based on such multimodal information.
Multimodal learning is especially important for robots that need to operate properly in the real world, because they need to make sense of the world based on the various types of information they receive through their onboard sensors~\cite{noda2014multimodal}.

\begin{figure}[tb]
\begin{center}
  \includegraphics[width=17cm]{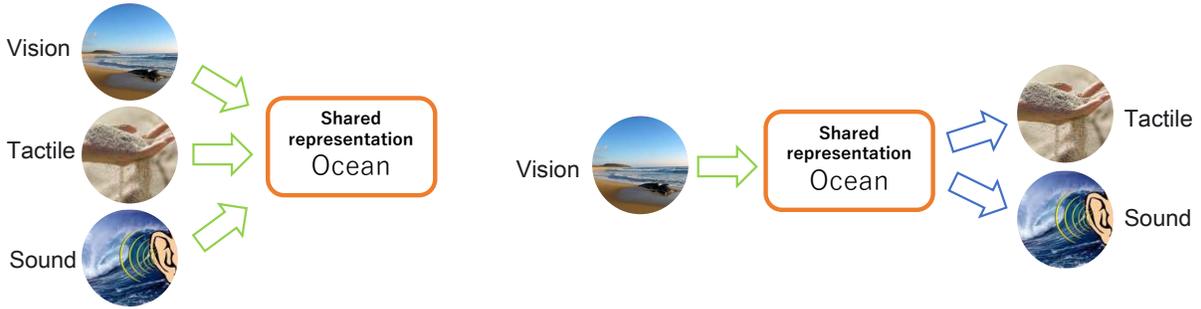} 
  \caption{Tasks in multimodal deep generative models. Left: Inference from different modalities to a shared representation. Right: Cross-modal generation.}
\label{fig:overview}
\end{center}
\end{figure}

The most fundamental challenge in multimodal learning is obtaining a compact and modality-invariant representation that integrates different modalities without label information, which we call a {\it shared representation}~\cite{ngiam2011multimodal}.
For example, we can construct an abstract shared representation of ``ocean'' in our brain by perceiving a view of sandy beach, the sound of waves, and the feeling of sand (Figure~\ref{fig:overview}, left). In learning a shared representation, different modalities are assumed to have complementarity for a given task, i.e., one modality contains information about a task that is unavailable in other modalities~\cite{lahat2015multimodal}. Under this assumption, learning shared representations is expected to lead to the acquisition of higher performance representations of different modalities. 

Another important challenge is to translate data between modalities, i.e., {\it cross-modal generation}.
This corresponds to the way we imagine the sound of waves and the feeling of sand when we look at a picture of a beach (Figure~\ref{fig:overview}, right).
If we have a way of embedding all modalities into a shared representation and generating modalities from it, then we can achieve generation between arbitrary modalities via this space. 

However, accomplishing these challenges requires that we consider the {\it heterogeneity} among different modalities. Heterogeneity indicates that they have different feature spaces and distributions~\cite{weiss2016survey}.
In the above example, the data type of a picture of the ocean is very different from that of the sound of waves, and the amount of information they contain for the concept of ``ocean'' is also very different.
Therefore, to deal with them properly, we must consider the non-deterministic relationships between them.

To address this issue, researchers have proposed a generative model-based approach that treats all multimodal data as stochastically generated from a global latent variable that corresponds to a shared representation~\cite{srivastava2012multimodal, blei2003modeling, nakamura2009grounding}.
The advantage of this is not only that multimodal data can be generated from a shared latent variable, but also that the latent variable can be inferred from arbitrary modalities, which allows the acquisition of shared representation and cross-modal generation.
However, conventional generative models cannot directly handle data with large dimensions, such as images, and require a large computational cost for inference.

Deep generative models~\cite{kingma2013auto, goodfellow2014generative, oord2016wavenet, van2016pixel, rezende2015variational, kingma2018glow} are a framework that represents the distribution of generative models by deep neural networks. 
These models are characterized by end-to-end learning with back-propagation and the ability to generate high-dimensional and complex data, thanks to deep neural networks.
In particular, variational autoencoders (VAEs)~\cite{kingma2013auto} enable fast inference to latent representations by parameterizing the inference to latent variables with neural networks, called amortized inference~\cite{gershman2014amortized}.
In this context, multimodal learning with deep generative models has attracted significant attention in recent years. 
In this paper, we refer to this research field as {\it multimodal deep generative models}.

This paper is a survey of multimodal deep generative models.
There have already been many surveys on multimodal (or multi-view) machine learning ~\cite{atrey2010multimodal, xu2013survey, sun2013survey, lahat2015multimodal, wang2016comprehensive, baltruvsaitis2018multimodal, gao2020survey, zhang2020multimodal}.
Among them, Baltru{\v{s}}aitis et  al.~\cite{baltruvsaitis2018multimodal} divide various multimodal learning studies into five challenges (representation, translation, alignment, fusion, and co-learning) and provide comprehensive descriptions of them.
However, Baltru{\v{s}}aitis et  al.~\cite{baltruvsaitis2018multimodal} do not fully cover the multimodal deep generative models that we survey in this paper; specifically, they do not include methods that use deep generative models in the ``representation'' task, and only describe conditional models in the ``translation'' task.
This paper is, to the best of our knowledge, the first comprehensive survey of multimodal deep generative models. Most of the deep generative models treated in this paper are based on VAEs, and we classify multimodal generative models into {\it coordinated} and {\it joint} models according to \cite{baltruvsaitis2018multimodal}.

For coordinated models, we focus on the definition of closeness between inference distributions in the objective and divide them into two criteria: the distance between inference distributions and cross-modal generation loss. In joint models, we classify  studies according to three main challenges: the handling of missing modalities, modality-specific latent variables, and weakly-supervised learning. The handling of missing modalities is classified into approaches that introduce surrogate unimodal inference and those that approximate joint inference by aggregating unimodal inference.

The rest of this paper is organized as follows.
Section~\ref{sec:2} describes the problem setting of multimodal generative models and the definitions of heterogeneity in multimodality and good shared representation.
Section~\ref{sec:3} gives an overview of VAEs, the advantages of using VAEs as multimodal generative models, and the categories of multimodal deep generative models.
Of these categories, Section~\ref{sec:4} describes coordinated models and Section~\ref{sec:5} describes joint models.
Section~\ref{sec:6} summarizes the benchmark datasets and applications in multimodal deep generative models.
Section~\ref{sec:7} discusses the future directions of multimodal deep generative models.
Section~\ref{sec:8} concludes the paper.

\section{Multimodal generative models}\label{sec:2}
\subsection{Notation}
Suppose that we are given an i.i.d. dataset $\mathbf{X} =\{X^{(i)}\}_{i=1}^N$, where each example $X^{(i)}=\{\mathbf{x}^{(i)}_{m}\}_{m=1}^M$ is a set of $M$ modalities and where each modality $\mathbf{x}^{(i)}_{m}=\{x^{(i)}_{1m},...,x^{(i)}_{D_mm}\} \in \mathcal{X}_{m}$ has its feature space $\mathcal{X}_{m}$.
We denote the true joint distribution of multimodal data as $p_{data}(X)=p_{data}(\mathbf{x}_1,...,\mathbf{x}_M)$.
And let $X_{k}^{(i)}$ denote a subset of $i$-th  multimodal data; that is, $X_{k}^{(i)} \subseteq	X^{(i)}$. We assume a representation $\mathbf{z}^{(i)}\in \mathcal{Z}$ that embeds different modalities $X^{(i)}$ and call it a shared representation.
In the following, $i$ might be omitted if the example is not specified for simplicity.

\subsection{Problem setting}
In this paper, we assume that the multimodal generative model is intended for both of the following two purposes (see also Figure 1): 
\begin{description}
    \item[1.] Embed all modalities $X$ in a good common space called the shared representation.
    \item[2.] Generate modalities $X_{k'}$\footnote{$X_{k'}$ means a subset of multimodal data different from $X_{k}$.} from arbitrary modalities $X_{k}$ via the shared representation $\mathbf{z}$.
\end{description}

The first purpose is sometimes referred to as multimodal representation learning~\cite{li2018survey}, and besides generative approaches, there are studies based on discriminative approaches such as contrastive learning~\cite{tian2019contrastive, alayrac2020self, tsai2021selfsupervised}. 

For the second purpose, it is often called cross-modal generation. There are mainstream studies based on conditional generative models that directly model $p(\mathbf{x}_1|\mathbf{x}_2)$, which is the transformation from one modality $\mathbf{x}_1$ to another $\mathbf{x}_2$~\cite{sohn2015learning, mirza2014conditional}. 
These models are also sometimes referred to as multimodal generative models~\cite{ivanovic2020multimodal}, but we do not include them in multimodal generative models in this paper because they do not acquire a shared representation and can only  generate cross-modally in one direction.

Multimodal generative models usually consist of latent variable models with all modalities $X$ as observed variables and the shared representation $\mathbf{z}$ as a latent variable. In most cases, each modality $\mathbf{x}_m$ is assumed to be conditionally independent given a latent variable; that is,
\begin{align}
    p_{\Theta}(X, \mathbf{z}) = \prod_{m=1}^{M} p_{\theta_m}(\mathbf{x}_m|\mathbf{z})p(\mathbf{z}) ,
\end{align}
where $\Theta=\{\theta_m\}_{m=1}^M$ is the set of parameters for the conditional distributions of each modality. 
When designing them, it is necessary to consider the {\it heterogeneity} of different modalities, which will be discussed in the next subsection.

The objective of multimodal generative models is the marginal log-likelihood given the data
and we aim to estimate the parameters that maximize it:
\begin{align}
    \hat{\Theta} = \arg\max_{\Theta} \mathbb{E}_{p_{data}(X)}\left[ \log p_{\Theta}(X) \right] = \arg\max_{\Theta} \mathbb{E}_{p_{data}(X)}\left[ \log  \int p_{\Theta}(X, \mathbf{z}) d\mathbf{z} \right].
\end{align}
However, this objective is intractable because it involves the marginalization of a latent variable.
Furthermore, to achieve the goal of a multimodal generative model, it is necessary to infer from the generative model, i.e., to find the posterior distribution $p_{\Theta}(\mathbf{z}|X_k)$ of the shared representation $\mathbf{z}$ given any modality $X_k$; however, the computation of this posterior is also intractable.

Therefore, the key issues for multimodal generative models are twofold: how to design and train the above generative models, and how to perform inference to a latent variable.
Before addressing these issues, we consider the heterogeneity in multimodality and the requirement for a good shared representation in the remaining subsections.

\subsection{Heterogeneity in multimodality}
In a multimodal learning framework, the term modality originally referred to each scheme and situation in which data are perceived~\cite{lahat2015multimodal, baltruvsaitis2018multimodal}, i.e., the process of obtaining the data. 
However, this term is more often seen as exclusively representing the type of information obtained by the process~\cite{ngiam2011multimodal}, a view this paper follows. 
Various types of modalities in our world--such as vision, sound, and smell--are said to be heterogeneous~\cite{shi2010transfer, weiss2016survey}. The term ``heterogeneity'' has been used in relation to the term ``domain'' in the study of transfer learning.

Then, what is the nature of heterogeneity in multimodal learning? To consider this, suppose that we are given two datasets, $\mathbf{X}_1=\{\mathbf{x}_{1}^{(i)}\}_{i=1}^N\in\mathcal{X}_1$ and $\mathbf{X}_2=\{\mathbf{x}_{2}^{(i)}\}_{i=1}^N\in\mathcal{X}_2$. In transfer learning, the domain of $\mathbf{X}_1$ is defined by a pairing of the data feature space $\mathcal{X}_1$ and the marginal distribution $p(\mathbf{X}_1)$~\cite{pan2009survey, weiss2016survey}. 
Therefore, the fact that the domains are different between the two datasets means either: 1. $\mathcal{X}_1=\mathcal{X}_2$ and $p(\mathbf{X}_1)\neq p(\mathbf{X}_2)$, 2. $\mathcal{X}_1\neq \mathcal{X}_2$ and $p(\mathbf{X}_1)= p(\mathbf{X}_2)$, or 3. $\mathcal{X}_1\neq\mathcal{X}_2$ and $p(\mathbf{X}_1)\neq p(\mathbf{X}_2)$. In the context of heterogeneous transfer learning, if the two datasets have different feature spaces, they are considered heterogeneous, i.e., either 2 or 3~\cite{weiss2016survey}. 

On the other hand, when considering heterogeneity in the multimodality, the information entropy differs between modalities and there is often no one-to-one invertible correspondence between them. In the example in Figure~\ref{fig:overview}, the view of a sandy beach is considered to contain the most information to represent the concept of an ocean. Furthermore, there are countless corresponding views of the same sound of waves or the feeling of sand. This consideration implies that different modalities have different distributions.

Therefore, heterogeneity in a multimodal context involves both differences, i.e., 3. For good multimodal learning, we need to address both differences when designing our model.

\subsection{Requirements for a good shared representation}
In general, a ``good representation'' is one that can be used for various tasks while retaining the characteristics of the original input~\cite{bengio2013representation}. Representation learning aims to learn this from data in an unsupervised manner. To achieve this objective, Bengio et al.~\cite{bengio2013representation} argued that we should consider general-purpose priors (or meta-priors~\cite{tschannen2018recent}) about the world around us as properties that such representations should have including disentanglement, manifolds, and hierarchical representation.

Srivastava et al.~\cite{srivastava2012multimodal} argued that a shared representation should further satisfy the following properties\footnote{
In the original text, a shared representation is called a {\it joint representation}; we have changed this terminology to match that in this paper. Strictly speaking, there is a subtle difference between shared representation and joint representation: the former refers to a space shared by different modalities while the latter refers to a shared representation of the {\it joint model} (see Section~\ref{sec:joint}), i.e., a space in which different modalities are fused.}:
\begin{description}
    \item[1.] The shared representation must be such that similarity in the representation space implies similarity of the corresponding ``concepts''. 
    \item[2.] The shared representation must be easy to obtain even in the absence of some modalities.
\end{description}

The first property requires that similar concepts cohere in a shared representation. 
The second implies that the representation should be inferred appropriately even from an input that is missing any modality, which is a necessary condition for a successful cross-modal generation.
In the following, we consider these properties as requirements for a good shared representation. 

\subsection{Related works of multimodal generative models before deep generative models}
\label{sec:related}
Even before the advent of deep generative models, various multimodal generative models have been proposed.

The most popular approaches are based on restricted Boltzmann machines (RBMs)~\cite{holyoak1987parallel,hinton2006reducing}. Xing et al.~\cite{xing2012mining} build a joint model of images and texts using a linear RBM. Srivastava et al.~\cite{srivastava2012learning} propose a multimodal generative model based on deep belief networks~\cite{hinton2006fast,hinton2009deep}, which has since been used in various multimodal  applications~\cite{kim2013deep, huang2013audio}.
Srivastava et al.~\cite{srivastava2012multimodal} propose to use  deep Boltzmann machines (DBMs)~\cite{salakhutdinov2009deep} for multimodal data, specifically images and text. Sohn et al.~\cite{sohn2014improved} introduce an improved method of multimodal DBM based on the idea of the variation of information.
Furthermore, multimodal DBMs are used in a variety of multimodal applications~\cite{ouyang2014multi, suk2014hierarchical, pang2015deep}.
The advantages of these models are that they facilitate inference from modalities to latent variables and thus cross-modal generation and that they can properly handle missing modalities of input. Furthermore, since these are probabilistic models, they can naturally take into account differences in the distributions of modalities. However, because these learning rules are based on the Markov chain Monte Carlo (MCMC) method, it is difficult to handle complex high-dimensional data directly as input; as a result, they cannot  handle large differences in features.

Other research involves extending topic models such as latent Dirichlet allocation (LDA)~\cite{blei2003latent} to multimodal input~\cite{blei2003modeling,nakamura2009grounding}.
Nakamura et al.~\cite{nakamura2009grounding} propose multimodal LDA to address the categorization of images, sounds, and haptic information in robots and the language grounding between them. The multimodal LDA has similar advantages as in the case of DBMs and various studies have extended this model~\cite{putthividhy2010topic, nakamura2011bag, araki2012online, zheng2014topic, nakamura2014mutual}; however, they are all based on topic models and cannot directly handle data with high dimensionality and complex structures such as images, thereby requiring preprocessing such as extracting ``visual words''.

In summary, there have been many generative model-based methods before deep generative models  and they have many advantages such as being able to handle differences in distributions and obtain good shared representations. However,they could not directly infer from or generate complex data such as images, which is a major drawback when dealing with heterogeneous multimodal information in practice.

\section{Multimodal deep generative models}\label{sec:3}
\subsection{Deep generative models and variational autoencoders}
Deep generative models are a group of generative models in which probability distributions are parameterized by deep neural networks.
They differ from conventional generative models in that they can directly handle high-dimensional data thanks to deep neural networks, and the entire model can be trained end-to-end using the gradient of neural networks by backpropagation. Deep generative models include variational autoencoders (VAEs)~\cite{kingma2013auto}, generative adversarial networks (GANs)~\cite{goodfellow2014generative}, autoregressive models~\cite{van2016pixel, oord2016wavenet}, flow-based models~\cite{rezende2015variational, kingma2018glow}, and so on.

VAEs are latent variable models of $p_{\theta}(\mathbf{x})=\int  p_{\theta}(\mathbf{x}|\mathbf{z})p(\mathbf{z}) d\mathbf{z}$, where $p_{\theta}(\mathbf{x}|\mathbf{z})$ is parameterized by a deep neural network and prior $p(\mathbf{z})$ is often set as a standard multivariate Gaussian. 
In VAEs, instead of directly maximizing the intractable marginal log-likelihood $\log p_{\theta}(\mathbf{x})$, we maximize its evidence lower bound (ELBO) given data $\mathbf{x}$:
\begin{eqnarray}
\mathcal{L}_{VAE}(\mathbf{x}) &\equiv& \mathbb{E}_{q_{\phi}(\mathbf{z}|\mathbf{x})}\left[ \log p_{\theta}(\mathbf{x}|\mathbf{z})\right] - D_{KL}[q_{\phi}(\mathbf{z}|\mathbf{x})||p(\mathbf{z})] \label{eq:elbo}\\
&=& \mathbb{E}_{q_{\phi}(\mathbf{z}|\mathbf{x})}\left[ \log \frac{p_{\theta}(\mathbf{x}|\mathbf{z})p(\mathbf{z})}{q_{\phi}(\mathbf{z}|\mathbf{x})}\right] \leq \log p_{\theta}(\mathbf{x}),
\end{eqnarray}
where $q_{\phi}(\mathbf{z}|\mathbf{x})$ is a posterior distribution that approximates inference $p_{\theta}(\mathbf{z}|\mathbf{x})$ and is also parameterized by the deep neural network. We designate $q_\phi(\mathbf{z}|\mathbf{x})$ as the {\it encoder} or {\it inference distribution}\footnote{In this paper, we use the term ``encoder'' to refer to any form of mapping from an input space to a latent space, whether deterministic or probabilistic, and  ``inference distribution'' to refer specifically to the conditional distribution $q_{\phi}(\mathbf{z}|\mathbf{x})$.} and $p_\theta(\mathbf{x}|\mathbf{z})$ as the {\it decoder}. Moreover, in Equation~\eqref{eq:elbo}, the first term represents a negative reconstruction loss of input $\mathbf{x}$, and the second term represents a regularization for the encoder. 
To obtain a good representation, especially a disentangled one~\cite{higgins2016beta}, the second term is often adjusted by introducing a coefficient, which is omitted in this paper for brevity.

\subsection{Advantages of VAEs as multimodal generative models}

Compared with existing multimodal generative models, normal deep autoencoders, and other deep generative models such as GANs, VAEs have various advantages in realizing the purpose of multimodal generative models.

First, VAEs represent both generation and inference as paths of DNNs; therefore, their training and execution are fast and they can handle high-dimensional and complex inputs. On the other hand, traditional DBM-based models take a long time to train and execute inference and cannot handle such inputs directly.

Second, VAEs are good models for representation learning. 
As mentioned earlier, good representation requires the inclusion of general-purpose priors in the model.
It is known that VAEs can easily incorporate such priors by adding constraints to the inference distribution and by explicitly assuming a graphical model structure~\cite{tschannen2018recent}.  In particular, the property of learning manifolds helps to realize the first requirement for good shared representations in Section 2.4. 
Normal deep autoencoders~\cite{vincent2010stacked} can also acquire a representation of the data by introducing bottleneck constraints and noise, which have been applied in multimodal settings~\cite{ngiam2011multimodal,jaques2017multimodal}; however, they cannot include various priors for a good representation, such as disentanglement and structural representation.

In addition, since VAEs are probabilistic models, they can explicitly represent differences in the distribution of data, unlike normal deep autoencoders.
In other words, VAEs can explicitly deal with the heterogeneity of multimodal data.

GANs are well-known as deep generative models other than VAEs, and various methods using multimodal information with GANs have been proposed~\cite{zhang2017stackgan,zhu2017unpaired, isola2017image,liu2017unsupervised, lee2018diverse}.
However, most of these are conditional models~\cite{zhang2017stackgan,isola2017image}, i.e., unidirectional, and even if they are bidirectional models, they do not deal with heterogeneous data (i.e., data with different distributions and dimensions)~\cite{zhu2017unpaired, liu2017unsupervised, lee2018diverse}.
In addition, in terms of representation learning, it is difficult for GANs to introduce general-purpose priors as flexibly as VAEs. 
Furthermore, the stability of their learning remains a challenge, such as mode collapse, which is confirmed to be more serious in a multimodal setting~\cite{wu2019multimodal}.
Therefore, VAEs have become the mainstream in multimodal deep generative models, where GANs are sometimes used to improve the quality of the generation of multimodal VAEs~\cite{suzuki2018improving,wu2019multimodal} or to implement divergence between distributions in VAEs~\cite{daunhawer2021self}.

\subsection{Categories of multimodal deep generative models}
\begin{figure}[tb]
\begin{center}
  \includegraphics[width=11cm]{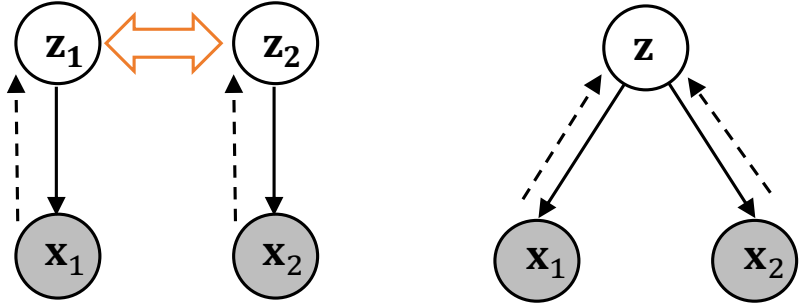} 
  \caption{Graphical models of two categories of multimodal deep generative models. In this figure, the number of modalities is assumed to be two. The gray circles represent the observed variables and the white ones represent the latent variables.  The black arrows represent the generating process of random variables, while the dotted ones represent the inference distributions. Furthermore, the orange two-way arrow indicates that the two latent variables are in the same space. Left: the coordinated model; Right: the joint model.}
\label{fig:joint_coordinated}
\end{center}
\end{figure}

We divide multimodal deep generative models into two major categories according to how they model inference to shared representations: modeling inference $q_{\phi_m}(\mathbf{z}_m|\mathbf{x}_m)$ from single modality $\mathbf{x}_m$, or modeling inference $q_{\Phi}(\mathbf{z}|X)$ from all modalities $X$.
According to Baltru{\v{s}}aitis et al.~\cite{baltruvsaitis2018multimodal}, the former is called the {\it coordinated} model and the latter is called the {\it joint} model. Figure~\ref{fig:joint_coordinated} shows graphical models of the two categories. 

Both of these categories aim to satisfy the two properties mentioned in Section 2.4. Many multimodal deep generative modeling methods focus mainly on how to achieve the second property since the first can be achieved through the representation learning of VAEs. However, the coordinated and joint models have slightly different goals for the second property: the coordinated model aims for the inference results from {\it each modality} to be the same, whereas the joint model aims for the inference results from {\it any set of modalities} to be the same.
We will discuss the study of the coordinated model in Section~\ref{sec:coordinated} and the joint model in Section~\ref{sec:joint}.

\begin{figure}[tb]
\begin{center}
  \includegraphics[width=16cm]{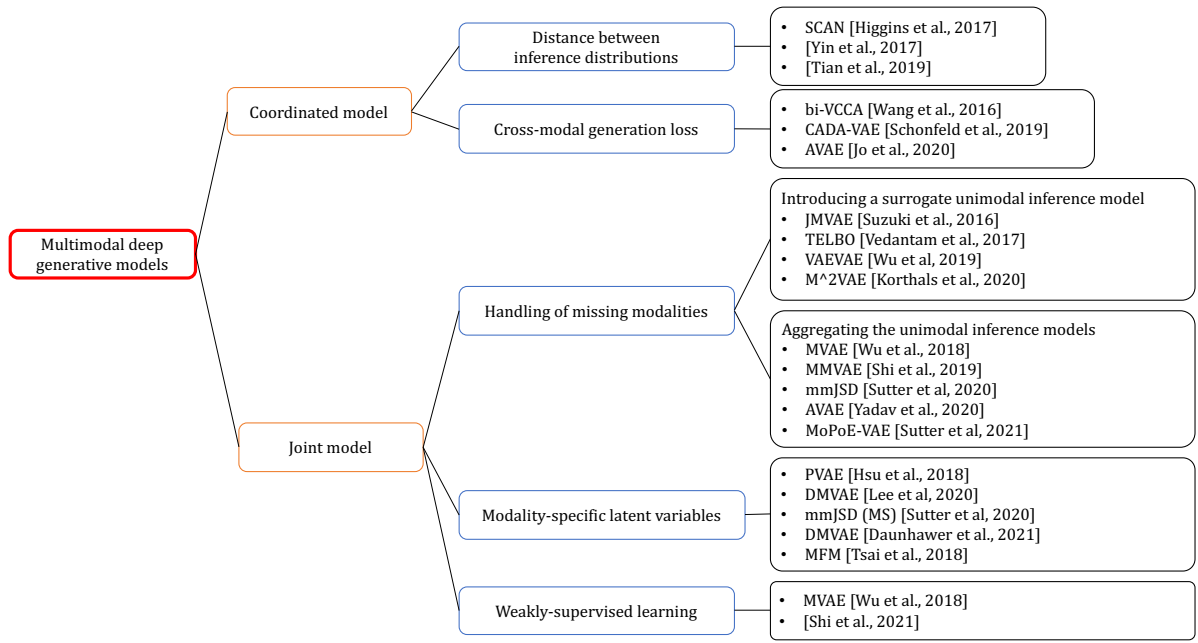} 
  \caption{A systematic diagram of the multimodal deep generative modeling methods and their categories described in this paper. ``mmJSD (MS)'' refers to being mmJSD~\cite{sutter2020multimodal} with modality-specific latent variables. Note that each category is not exclusive. For example, MVAE~\cite{wu2018multimodal} is included in both the ``handling of missing modalities'' and ``weakly-supervised learning'' categories. In addition, all the methods in the categories ``modality-specific latent variables'' and ``weakly-supervised learning'' address the missing modality problem but their main focus is different; therefore, they are treated as different categories. JVAE~\cite{suzuki2016joint,vedantam2017generative} is not included here because it is the simplest of the joint models and does not address any of the difficulties corresponding to the categories.}
\label{fig:chart}
\end{center}
\end{figure}

See Figure~\ref{fig:chart} for a systematic view of multimodal deep generative models and their categories; see Table~\ref{tab:list} for the differences in their properties.

\setcounter{page}{1}
\begin{table}[tb]
\caption{List of multimodal deep generative models and their properties. 
``Modality-specific'' refers to whether a model contains modality-specific latent variables, ``end-to-end'' refers to whether a model does end-to-end learning with a single objective function, and ``scalability'' refers to whether the cost of the model increases exponentially with the number of modalities.
Note that models marked ``-'' in the scalability column mean that they are coordinate models and do not assume more than two modalities.}
\tiny
\centering
  \begin{tabular}{l|lllll}
Models                                                    & Aggregated inference & Missing modalities                & Modality-specific & End-to-end & Scalability      \\\hline
SCAN~\cite{higgins2017scan}                               & $\times$          & \checkmark (single modality) & $\times$                             & $\times$ & -                                                                   \\
\cite{yin2017associate}                                   & $\times$          & \checkmark (single modality) & $\times$                             & \checkmark & -                                       \\
\cite{tian2019latent}                                     & $\times$          & \checkmark (single modality) & \checkmark                         & \checkmark & -                                       \\
bi-VCCA~\cite{wang2016deep}                               & $\times$          & \checkmark (single modality) & $\times$                             & \checkmark & -                                       \\
CADA-VAE~\cite{schonfeld2019generalized}                  & $\times$          & \checkmark (single modality) & $\times$                             & \checkmark & -                                       \\
AVAE~\cite{jo2019cross}                                   & $\times$          & \checkmark (single modality) & $\times$                             & $\times$ & \checkmark                              \\
JVAE       & \checkmark      & $\times$                            & $\times$                             & \checkmark & \checkmark                              \\
JMVAE~\cite{suzuki2016joint}                              & \checkmark      & \checkmark                        & $\times$                             & \checkmark & $\times$ (memory cost)        \\
TELBO~\cite{vedantam2017generative}                       & \checkmark      & \checkmark                        & $\times$                             & $\times$ & $\times$ (memory cost)        \\
VAEVAE~\cite{wu2018multimodal}                            & \checkmark      & \checkmark                        & $\times$                             & \checkmark & $\times$ (memory cost)        \\
$M^2$VAE~\cite{korthals2019multi}                           & \checkmark      & \checkmark                        & $\times$                             & \checkmark & $\times$ (memory cost)        \\
MVAE~\cite{wu2018multimodal}                              & \checkmark      & (\checkmark) (require sub-sampling)             & $\times$                             & \checkmark & \checkmark       \\
MMVAE~\cite{shi2019variational}                           & $\times$ (MoE)    & \checkmark                        & $\times$                             & \checkmark & \checkmark                              \\
mmJSD~\cite{sutter2020multimodal}                         & $\times$ (MoE)    & \checkmark                        & $\times$                             & \checkmark & \checkmark                              \\
mmJSD (MS)~\cite{sutter2020multimodal} & $\times$ (MoE)    & \checkmark                        & \checkmark                         & \checkmark & \checkmark                              \\
AVAE~\cite{yadav2020bridged}                              & \checkmark      & \checkmark                        & $\times$                             & $\times$ & \checkmark                              \\
MoPoE-VAE~\cite{sutter2021generalized}                    & \checkmark      & \checkmark                        & $\times$                             & \checkmark & $\times$ (computational cost) \\
PVAE~\cite{hsu2018disentangling}                          & \checkmark      & \checkmark                        & \checkmark                         & \checkmark & $\times$ (memory cost)        \\
DMVAE~\cite{lee2020private}                               & \checkmark      & \checkmark                        & \checkmark                         & \checkmark & \checkmark                              \\
DMVAE~\cite{daunhawer2021self}                             & \checkmark      & \checkmark                        & \checkmark                         & \checkmark & \checkmark                              \\
MFM~\cite{tsai2018learning}                          & \checkmark      & \checkmark                        & \checkmark                         &  \checkmark & $\times$ (memory cost)        \\
\cite{shi2021relating}                                    & $\times$ (MoE)    & \checkmark                        & $\times$                              & \checkmark & \checkmark
	\label{tab:list}
  \end{tabular}
\end{table}

\section{Coordinated model}\label{sec:4}
\label{sec:coordinated}
The goal of the coordinated model is to bring the inference distributions conditioned on the different modalities closer together\footnote{Note that the inference distributions are the same when the outputs of the encoder networks (i.e., the parameters of the inference distributions) are the same, not necessarily when the values of the training parameters of these networks are all the same.}. This leads to making arbitrary representations of different modalities the same; that is, 
$\mathbf{z}_1^{(i)}\approx\mathbf{z}_2^{(i)} \forall i\in \{1,...,N\}$, where $\mathbf{z}_1^{(i)}\sim q_{\phi_1}(\mathbf{z}_1|\mathbf{x}^{(i)}_1)$ and $ \mathbf{z}_2^{(i)} \sim q_{\phi_2}(\mathbf{z}_2|\mathbf{x}^{(i)}_2).$
Correspondingly, we consider generated models for each modality, i.e., $p_{\theta_1}(\mathbf{x}_1)=\int p_{\theta_1}(\mathbf{x}_1|\mathbf{z}_1)p(\mathbf{z}_1)d\mathbf{z}_1$ and $p_{\theta_2}(\mathbf{x}_2)=\int p_{\theta_2}(\mathbf{x}_2|\mathbf{z}_2)p(\mathbf{z}_2)d\mathbf{z}_2$
,  instead of the joint generative model $p_{\Phi}(\mathbf{x}_1,\mathbf{x}_2)=\int p_{\theta_1}(\mathbf{x}_1|\mathbf{z})p_{\theta_2}(\mathbf{x}_2|\mathbf{z})p(\mathbf{z})d\mathbf{z}$ (see Figure~\ref{fig:joint_coordinated}).

The objective of VAEs in the coordinated version consists of the ELBOs for all modalities, $\mathbf{x}_1$ and $\mathbf{x}_2$, and a loss function that represents the closeness between inference distributions of each modality $\mathcal{L}_{coor}(\mathbf{x}_1, \mathbf{x}_2)$; that is,
\begin{equation}
    \left(\sum_{\mathbf{x}_m\in \{\mathbf{x}_1, \mathbf{x}_2\}}\mathcal{L}_{VAE}(\mathbf{x}_m)\right) - \mathcal{L}_{coor}(\mathbf{x}_1, \mathbf{x}_2).
\end{equation}
In many cases, the term $\mathcal{L}_{coor}(\mathbf{x}_1, \mathbf{x}_2)$ is multiplied by a weight factor, which is omitted in this paper for brevity. This is also the case for the equations in the following studies.
Whether the term $\mathcal{L}_{coor}$ is learned simultaneously with ELBOs or separately depends on each method.
In this section, the number of modalities is assumed to be two.

\begin{figure}[tb]
\begin{center}
  \includegraphics[width=13cm]{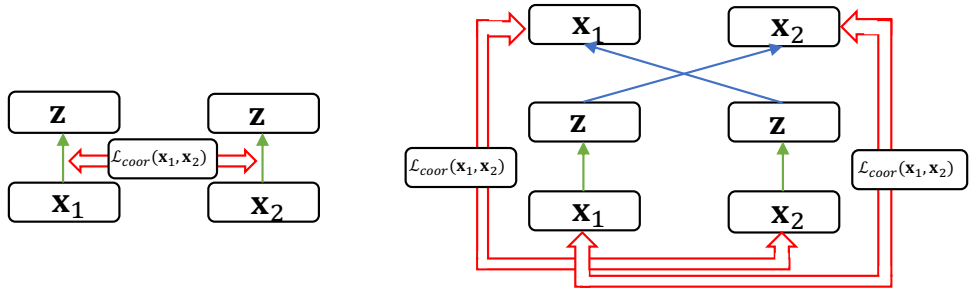} 
  \caption{Difference in the calculation method of $\mathcal{L}_{coor}$ (represented by red two-way arrows) in coordinated models. Left: distance between inference distributions; Right: cross-modal generation loss.}
\label{fig:coordinated}
\end{center}
\end{figure}

There are several ways to define $\mathcal{L}_{coor}$, but in this paper we will focus on the following two criteria: {\it the distance between inference distributions} and {\it cross-modal generation loss} (see Figure~\ref{fig:coordinated}).
In addition to these, cycle-consistency loss is used in image-to-image translation, which is the task of transforming an image belonging to one domain to another domain~\cite{zhu2017unpaired, isola2017image}. Although some of these models consider shared latent variables in different domains~\cite{liu2017unsupervised, lee2018diverse}, they assume that the input is two domain images of the same size, not heterogeneous modalities; therefore, we do not include them in this survey.

The characteristics of the models in this category are that they are independently proposed in different research areas such as zero-shot learning, domain adaptation, and symbol grounding. Therefore, each model is often compared with previous studies in each area and in each problem setting.

\subsection{Distance between inference distributions}

The most intuitive loss function for $\mathcal{L}_{coor}$ is the one that takes the distance between inferences of different modalities. 

In Yin et al.~\cite{yin2017associate}, the distance between inference distributions is the Kullback–Leibler (KL) divergence in both directions, which is learned simultaneously with the generative models:
\begin{equation}
    \mathcal{L}_{coor}(\mathbf{x}_1, \mathbf{x}_2) = D_{KL}(q_{\phi_1}(\mathbf{z}_1|\mathbf{x}_1)||q_{\phi_2}(\mathbf{z}_2|\mathbf{x}_2)) + D_{KL}(q_{\phi_2}(\mathbf{z}_2|\mathbf{x}_2)||q_{\phi_1}(\mathbf{z}_1|\mathbf{x}_1)).
\end{equation}

Symbol-Concept Association Network (SCAN)~\cite{higgins2017scan}, on the other hand, assumes a single KL divergence:
\begin{equation}
    \mathcal{L}_{coor}(\mathbf{x}_1, \mathbf{x}_2) = D_{KL}(q_{\phi_1}(\mathbf{z}_1|\mathbf{x}_1)||q_{\phi_2}(\mathbf{z}_2|\mathbf{x}_2)),
\end{equation}
where $\mathbf{x}_1$ is the image modality and $\mathbf{x}_2$ is the symbol one. Higgins et al.~\cite{higgins2017scan} select this direction so that the inference of the symbol covers the whole inference of the image.
Also, unlike Yin et al.~\cite{yin2017associate}, they first learn the VAE of $\mathbf{x}_1$, then fix this parameter and optimize the ELBO on $\mathbf{x}_2$ and $\mathcal{L}_{coor}(\mathbf{x}_1, \mathbf{x}_2)$.

Tian et al.~\cite{tian2019latent} consider the translation of different modalities as a domain transfer. 
To mitigate the heterogeneity of different modalities, they learn VAEs (or GAN) for each modality and then infer a representation $\mathbf{z}_m$ for each modality. Then, they take each of them as input and use an encoder $q(\mathbf{z}'|\mathbf{z}_m,m)$ that maps to a shared representation $\mathbf{z}_m'$ conditioned on the variable $m\in \{1,2\}$ representing the type of modality, and decode the shared representation into each representation with $p(\mathbf{z}_m|\mathbf{z}',m)$. 
They optimize the encoder and decoder in the VAE framework while simultaneously optimizing the sliced Wasserstein distance~\cite{bonneel2015sliced} between the embeddings in each modality, $\mathbf{z}'_1$ and $\mathbf{z}'_2$, and the loss in class label prediction from each embedding.

\subsection{Cross-modal generation loss}
Wang et al.~\cite{wang2016deep} propose a variational canonical correlation analysis (CCA) that extends linear CCA~\cite{hotelling1992relations} to deep probabilistic latent variable models, which can be regarded as changing the encoder of the joint VAE (described in Section 5) as $q_{\phi_1}(\mathbf{z}|\mathbf{x}_1)$.
Furthermore, they introduce an objective called bi-VCCA to learn the encoder of $\mathbf{x}_2$. This objective can be considered that $\mathcal{L}_{coor}$ in the coordinated model is set as follows:
\begin{equation}
    \mathcal{L}_{coor}(\mathbf{x}_1, \mathbf{x}_2) = - \mathbb{E}_{q_{\phi_1}(\mathbf{z}|\mathbf{x}_1)}\left[ \log p_{\theta_2}(\mathbf{x}_2|\mathbf{z})\right] - \mathbb{E}_{q_{\phi_2}(\mathbf{z}|\mathbf{x}_2)}\left[ \log p_{\theta_1}(\mathbf{x}_1|\mathbf{z})\right].
\end{equation}
This is the cross-modal generation loss between each modality, and it encourages encoding from any modality to generate both modalities by optimizing simultaneously with the ELBOs for all modalities: $\sum_{\mathbf{x}_m} \mathcal{L}_{VAE}(\mathbf{x}_m)$.

Cross and distribution aligned VAE (CADA-VAE)~\cite{schonfeld2019generalized} is learned to minimize both the distance between inference distributions and the cross-modal generation loss:
\begin{equation}
\mathcal{L}_{coor}(\mathbf{x}_1, \mathbf{x}_2) =
    \left(\left\|\boldsymbol{\mu}_1-\boldsymbol{\mu}_2\right\|_{2}^{2}+\left\|\boldsymbol{\Sigma}_1^{\frac{1}{2}}-\boldsymbol{\Sigma}_2^{\frac{1}{2}}\right\|_{F}^{2}\right)^{\frac{1}{2}} + \left(\left|\mathbf{x}_1-D_1\left(E_2\left(\mathbf{x}_2\right)\right)\right| + \left|\mathbf{x}_2-D_2\left(E_1\left(\mathbf{x}_1\right)\right)\right|\right),
\end{equation}
where $\boldsymbol{\mu}_m$ and $\boldsymbol{\Sigma}_{m}$ are the mean vector and covariance matrix of the Gaussian inference $q_{\phi_m}(\mathbf{z}|\mathbf{x}_m)$, and $||\cdot||_{F}$ is the Frobenius norm. Also, $E_m$ is the deterministic encoder function of the VAE for the modality $\mathbf{x}_m$, and $D_m$ is its deterministic decoder function.
The first term is the distance between inference distributions by the 2-Wasserstein distance\footnote{The 2-Wasserstein distance is the $p$-Wasserstein distance of order $p = 2$.}, and the second term corresponds to the cross-modal generation loss.
Schonfeld et al.~\cite{schonfeld2019generalized} propose CADA-VAE for generalized zero-shot learning~\cite{xian2017zero,pourpanah2020review}, which is used to embed different modalities, namely image features and class attributes, into a shared representation.

Jo et al.~\cite{jo2020associative} propose an associative variational autoencoder (AVAE), which has associators,  $p_{\rho_{21}}(\mathbf{z}_2|\mathbf{z}_1)$ and $p_{\rho_{12}}(\mathbf{z}_1|\mathbf{z}_2)$, to map between latent variables of different modalities. Associators are named after human associative learning~\cite{bliss1993synaptic}.
Using these associators, the inference from one modality $\mathbf{x}_1$ to the latent variable $\mathbf{z}_2$ of another modality $\mathbf{x}_2$ becomes $q_{\rho_{21},\phi_{2}}(\mathbf{z}_2|\mathbf{x}_1)=\int p_{\rho_{ji}}(\mathbf{z}_2|\mathbf{z}_1)q_{\phi_{i}}(\mathbf{z}_1|\mathbf{x}_1)d\mathbf{z}_1$.
After learning the VAE for each modality, they learn the associators by minimizing the following loss function:
\begin{equation}
   \mathcal{L}_{coor}(\mathbf{x}_1,\mathbf{x}_2) = \mathcal{L}_{assoc}( \mathbf{x}_1,\mathbf{x}_2) + \mathcal{L}_{assoc}(\mathbf{x}_2,\mathbf{x}_1),
\end{equation}
where $\mathcal{L}_{assoc}$ consists of a cross-modal generation loss and a regularization term for inference as follows:
\begin{equation}
\mathcal{L}_{assoc}( \mathbf{x}_i,\mathbf{x}_j)=-\mathbb{E}_{q_{\rho_{ji},\phi_{i}}(\mathbf{z}_j|\mathbf{x}_i)}\left[ \log p_{\theta_j}(\mathbf{x}_j|\mathbf{z}_j)\right] +  D_{KL}[q_{\rho_{ji},\phi_{i}}(\mathbf{z}_j|\mathbf{x}_i)||p(\mathbf{z}_j)].
\end{equation}

\section{Joint model}\label{sec:5}
\label{sec:joint}
The coordinated model can embed each modality into the same shared representation using the encoder for each modality.
However, this model cannot perform inference from all modalities.
The joint model, on the other hand, directly models the inference $q_{\Phi}(\mathbf{z}|X)$ to the shared latent space given all modalities $X=\{\mathbf{x}_m\}_{m=1}^M$.

An early joint model in deep neural networks was proposed by Ngiam et al.~\cite{ngiam2011multimodal}, which pre-trains autoencoders for each modality, then fine-tunes the mapping from each latent variable to the shared space that fuses them.
Wang et al.~\cite{wang2015deep} propose to pre-train the multimodal autoencoders with DBMs. However, as mentioned above, these models are insufficient for large multimodal datasets.

In the joint model with VAEs, the inference distribution $q_{\Phi}(\mathbf{z}|X)$ and the generative model $p_{\Theta}(X)=\int p(\mathbf{z})p_{\Theta}(X|\mathbf{z})d\mathbf{z}=\int p(\mathbf{z}) \prod_{\mathbf{x}_m \in X}p_{\theta_m}(\mathbf{x}_m|\mathbf{z})d\mathbf{z}$ are trained to optimize the following objective:
\begin{eqnarray}
\mathcal{L}_{JVAE}(X) &=& \mathbb{E}_{q_{\Phi}(\mathbf{z}|X)}\left[ \log p_{\Theta}(X|\mathbf{z})\right] - D_{KL}[q_{\Phi}(\mathbf{z}|X)||p(\mathbf{z})]\nonumber\\
&=& \mathbb{E}_{q_{\Phi}(\mathbf{z}|X)}\left[\sum_{\mathbf{x}_m\in X} \log p_{\Theta}(\mathbf{x}_m|\mathbf{z})\right] - D_{KL}[q_{\Phi}(\mathbf{z}|X)||p(\mathbf{z})],
\label{eq:jvae}
\end{eqnarray}
which is an extension of the VAEs' objective input to multiple modalities. In this paper, we refer to this as a joint VAE (JVAE) in accordance with Vedantam et al.~\cite{vedantam2017generative}.
JVAE is almost the same as the joint modeling by autoencoders~\cite{ngiam2011multimodal,wang2015deep}, but differs in that it uses probabilistic deep latent variable models.

JVAE has three major challenges: handling of missing modalities, modality-specific latent variables, and weakly-supervised learning. The following subsections describe the studies surrounding each challenge.

Each model in this category is closely related to each other, unlike the category of the coordinated model. First, Suzuki et al.~\cite{suzuki2016joint} proposed JVAE and how to handle missing modalities, then methods with other objective~\cite{vedantam2017generative, korthals2019multi, wu2019multimodal, sutter2020multimodal}, such as Vedantam et al.~\cite{vedantam2017generative}, and more effective methods to handle missing modalities~\cite{wu2018multimodal, shi2019variational, yadav2020bridged, sutter2021generalized}, such as Wu et al.~\cite{wu2018multimodal}, were proposed.
Wu et al.~\cite{wu2018multimodal} also proposed a method for a weakly-supervised learning setting where all modalities are not present at training, followed by Shi et al.~\cite{shi2021relating}.
Modality-specific latent variables were first introduced into JVAE by Hsu et al.~\cite{hsu2018disentangling} and Tsai et al.~\cite{tsai2018learning}, and then models with various inference methods and objectives were proposed~\cite{sutter2020multimodal, lee2020private, daunhawer2021self}.

\subsection{Handling of missing modalities}
Cross-modal generation is one of the necessary tasks in multimodal generative models.
After training JVAE with Equation~\eqref{eq:jvae}, cross-modal generation  is performed as
\begin{equation}
p(X_{k'}|X_{k})=\int q_{\Phi}(\mathbf{z}|X_{k})\prod_{\mathbf{x}_m\in X_{k'}}p_{\theta_m}(\mathbf{x}_m|\mathbf{z})d\mathbf{z}.
\end{equation}
Whether this can be done properly depends on whether we can infer from an arbitrary set of modalities with the inference distribution $q_{\Phi}$, i.e., whether we can handle missing input modalities, which is also one of the requirements for a good shared representation.
In multimodal generative models prior to deep neural networks, inference is based on sampling methods such as MCMC, which can deal with this missing issue.
However, JVAE cannot deal with missing modalities, because inference is performed by forward propagation of a deep neural network that parameterizes the inference distribution. Suzuki et al.~\cite{suzuki2018improving} show that when informative modality inputs (e.g., images) are missing from the input of a neural network-based inference distribution, the inferred representation is significantly corrupted.
Therefore, how to handle missing modalities is one of the most important and challenging issues in JVAE.

\begin{figure}[tb]
\begin{center}
  \includegraphics[width=13cm]{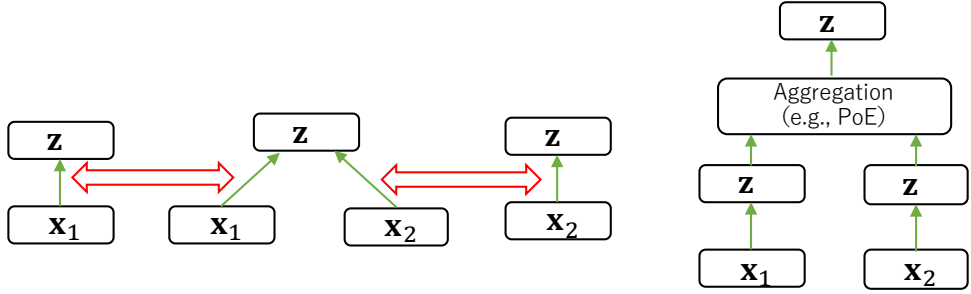} 
  \caption{Two different approaches to handling missing modalities. Left: Introducing surrogate unimodal inference distributions. The red two-way arrows mean that each surrogate model approximates a joint inference distribution. Right: Approximating a joint inference distribution by aggregating unimodal inference distributions with  PoE, MoE, etc.}
\label{fig:joint}
\end{center}
\end{figure}

There are two main approaches to deal with the above issue. One is to introduce a {\it surrogate unimodal inference distribution}, and the other is to approximate the joint inference distribution by {\it aggregating the unimodal inference distributions} (see Figure~\ref{fig:joint}).

\subsubsection{Introducing a surrogate unimodal inference distribution}

Suzuki et al.~\cite{suzuki2016joint} introduce surrogate unimodal inference distributions, $q_{\phi_1}(\mathbf{z}|\mathbf{x}_1)$ and $q_{\phi_2}(\mathbf{z}|\mathbf{x}_2)$, and learn these and the joint model simultaneously by optimizing the following objective:
\begin{equation}
\mathcal{L}_{JVAE}(\mathbf{x}_1, \mathbf{x}_2) +  D_{KL}[q_{\Phi}(\mathbf{z}|\mathbf{x}_1,\mathbf{x}_2)||q_{\phi_1}(\mathbf{z}|\mathbf{x}_1)] + D_{KL}[q_{\Phi}(\mathbf{z}|\mathbf{x}_1,\mathbf{x}_2)||q_{\phi_2}(\mathbf{z}|\mathbf{x}_2)],
\end{equation}
which is referred to as a joint multimodal VAE (JMVAE)\footnote{In the original paper~\cite{suzuki2016joint}, JVAE and JMVAE are referred to as JMVAE and JMVAE-kl, respectively. However, because some papers refer to JMVAE-kl as JMVAE~\cite{vedantam2017generative, wu2018multimodal}, we follow them to avoid confusion in terminology.}. 
These additional KL divergence terms encourage each unimodal inference distribution to approximate the joint inference distribution of JVAE.
Suzuki et al.~\cite{suzuki2016joint} prove that this objective is a lower bound on the variation of information
\begin{equation}
-\mathbb{E}_{p_{data}(\mathbf{x}_1, \mathbf{x}_2)}[\log p(\mathbf{x}_1|\mathbf{x}_2)+\log p(\mathbf{x}_2|\mathbf{x}_1)].
\end{equation}
That is,  JMVAE is optimized to encourage cross-modal generation.
Also, Vedantam et al.~\cite{vedantam2017generative} prove that the lower bound of the expectation of the above added KL divergence term is equal the KL divergence between unimodal inference distribution and  ``averaged'' distribution of joint inference distribution 
$q^{avg}(\mathbf{z}|\mathbf{x}_1)=\int q_{\Phi}(\mathbf{z}|\mathbf{x}_1, \mathbf{x}_2) p_{data}(\mathbf{x}_2|\mathbf{x}_1) d{\mathbf{x}_2}$; that is,
\begin{equation}
\mathbb{E}_{p_{data}(\mathbf{x}_1, \mathbf{x}_2)}[D_{KL}[q_{\Phi}(\mathbf{z}|\mathbf{x}_1,\mathbf{x}_2)||q_{\phi_1}(\mathbf{z}|\mathbf{x}_1)]]\geq D_{KL}[q^{avg}_{\Phi}(\mathbf{z}|\mathbf{x}_1)||q_{\phi_1}(\mathbf{z}|\mathbf{x}_1)].
\end{equation}

Vedantam et al.~\cite{vedantam2017generative} introduce the following objective to learn unimodal inference distributions:
\begin{equation}
\mathcal{L}_{JVAE}(\mathbf{x}_1, \mathbf{x}_2) + \mathcal{L}_{VAE}(\mathbf{x}_1) + \mathcal{L}_{VAE}(\mathbf{x}_2),
\label{eq:telbo}
\end{equation}
which consists of ELBOs for each given all combinations of all modalities. 
In the case of two modalities, there are three ELBOs; therefore, this objective is called triple ELBO (TELBO).
In addition, while JMVAE is learned end-to-end, in TELBO, the joint model is trained first, followed by the unimodal inference distributions.

Korthals et al.~\cite{korthals2019multi} propose a method to learn unimodal inference distributions by the following objective:
\begin{equation}
\mathcal{L}_{JVAE}(\mathbf{x}_1, \mathbf{x}_2) + \mathcal{L}_{VAE}(\mathbf{x}_1) + \mathcal{L}_{VAE}(\mathbf{x}_2) + D_{KL}[q_{\Phi}(\mathbf{z}|\mathbf{x}_1, \mathbf{x}_2)||q_{\phi_1}(\mathbf{z}|\mathbf{x}_1)] + D_{KL}[q_{\Phi}(\mathbf{z}|\mathbf{x}_1, \mathbf{x}_2)||q_{\phi_2}(\mathbf{z}|\mathbf{x}_2)],
\end{equation}
which can be viewed as a combination of JMVAE and TELBO's lower bounds and is called a multi-modal VAE (M$^2$VAE). Korthals et al.~\cite{korthals2019multi} show that this objective can be derived as a lower bound on the joint log-likelihood by a factor of two: $2\log p_{\Theta}(\mathbf{x}_1, \mathbf{x}_2)$.

By a derivation similar to M$^2$VAE, Wu et al.~\cite{wu2019multimodal} propose VAEVAE; the difference with M$^2$VAE is that VAEVAE does not have a KL divergence term between joint inference and prior.

These approaches are problematic in terms of memory and computational cost because they require additional inference distributions. In particular, when the number of modalities increases to three or more, the inference distribution must be prepared for all possible combinations of modalities, which results in an exponential increase in the size of the network.

\subsubsection{Aggregating the unimodal inference distributions}

An effective way to solve the scale problem in surrogate unimodal inference is to approximate a joint inference by aggregating unimodal inference distributions with an arbitrary function $f$; that is,
\begin{equation}
q(\mathbf{z}|X) = f(\{q_{\phi_m}(\mathbf{z}|\mathbf{x}_m)\}_{m=1}^M).
\label{eq:poe}
\end{equation}
This approximation has advantages in terms of memory and computational cost because the number of networks increases only linearly with the number of modalities.

\begin{figure*}[tb]
 		\begin{center}
		\includegraphics[scale=0.8]{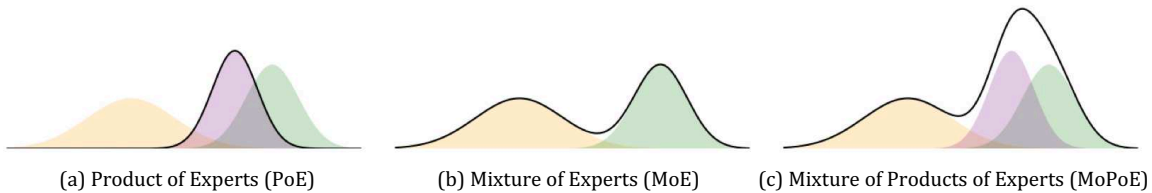}	
		\end{center}
			\caption{Methods that aggregate the inference distributions of each modality (yellow and green). Black lines represent the probability density of the aggregated joint distribution. Purple is the aggregated inference distribution using PoE. Here, the distribution of MoPoE is not normalized for the sake of clarity.}
		\label{fig:approaches}
\label{fig:aggregated}
\end{figure*}

Wu et al.~\cite{wu2018multimodal} consider unimodal inference as ``expert'' and propose approximating joint inference using the following product-of-experts (PoE)~\cite{hinton2002training} form:
\begin{equation}
q_{PoE}(\mathbf{z}|X) \propto p(\mathbf{z})\prod_{\mathbf{x}_m\in X} q_{\phi_m}(\mathbf{z}|\mathbf{x}_m).
\label{eq:poe}
\end{equation}
Usually, this posterior distribution cannot be calculated in closed form, but it can be calculated by constraining the unimodal inference and prior to a Gaussian distribution.
If no modalities are observed, this posterior is equal to prior. Then, as the number of modalities increases, the precision increases due to the product property—that is, this posterior becomes sharper (Figure~\ref{fig:aggregated}(a)).

Multimodal VAE (MVAE), proposed by Wu et al.~\cite{wu2018multimodal}, approximates the inference distribution by Equation~\eqref{eq:poe} and is learned by optimizing Eq~\eqref{eq:telbo}—i.e., ELBOs for each given all combinations of all modalities.
However, if more than three modalities are given, the objective has exponentially more terms; therefore, Wu et al.~\cite{wu2018multimodal} propose that the objective be a sub-sampling of ELBOs in any combination of modalities.
Also, Wu et al.~\cite{wu2019multimodal} propose training the inference distribution of PoE with the VAEVAE objective.

The shortcoming in approximating inference distributions with PoE is that, if the inference for a particular modality is very sharp, the joint inference will be heavily dominated by it; therefore, the optimization of unimodal inference with low precision might be greatly degraded.

Another way to approximate the joint inference is to use the mixture of experts (MoE) form—i.e., representing the joint inference distributions by sum of unimodal inferences as follows:
\begin{equation}
q_{MoE}(\mathbf{z}|X) = \sum_{\mathbf{x}_m \in X} \alpha_m \cdot q_{\phi_m}(\mathbf{z}|\mathbf{x}_m),
\end{equation}
where $a_m$ is constrained to $\sum_m a_m=1$, and in many cases $\alpha_m=\frac{1}{M}$.
Since MoE is the sum of each expert, the posterior distribution is not dominated by experts with high precision as in PoE, but spreads its density over all individual experts, evenly in the case of $\alpha_m=\frac{1}{M}$ (see Figure~\ref{fig:aggregated}(b)). Thus, individual experts can be optimized appropriately.

Shi et al.~\cite{shi2019variational} propose an MoE multimodal variational autoencoder (MMVAE) in which the joint inference distribution is approximated in the form of MoE.
The objective of MMVAE is the same as that of JVAE, which is as follows:
\begin{align}
&\mathbb{E}_{q_{\Phi}(\mathbf{z}|X)}\left[\sum_{\mathbf{x}_m\in X} \log p_{\Theta}(\mathbf{x}_m|\mathbf{z})\right] - D_{KL}[q_{\Phi}(\mathbf{z}|X)||p(\mathbf{z})]\nonumber\\
&=\frac{1}{M} \sum_{m=1}^{M} \mathbb{E}_{q_{\phi_{m}}\left(\mathbf{z} \mid \mathbf{x}_{m}\right)}\left[\log  p_{\Theta}\left( X|\mathbf{z}\right)\right] - D_{KL}\left(\frac{1}{M} \sum_{m=1}^{M}  q_{\phi_{m}}\left(\mathbf{z} \mid \mathbf{x}_{m}\right) || p(\mathbf{z})\right).
\end{align}
The first term corresponds to encouraging cross-modal generation.
Shi et al.~\cite{shi2019variational} apply the importance weighted autoencoder (IWAE)~\cite{burda2015importance} to make this ELBO tighter.

Sutter et al.~\cite{sutter2020multimodal} propose multimodal JS-divergence (mmJSD), which uses Jensen–Shannon (JS) divergence instead of KL divergence in the MMVAE objective:

\begin{equation}
\frac{1}{M} \sum_{m=1}^{M} \mathbb{E}_{q_{\phi_{m}}\left(\mathbf{z} \mid \mathbf{x}_{m}\right)}\left[\log  p_{\Theta}\left( X|\mathbf{z}\right)\right] - D_{JS_{\pi}^{M+1}}\left(\left\{q_{\phi_{m}}\left(\mathbf{z} \mid \mathbf{x}_{m}\right)\right\}_{m=1}^{M}, p(\mathbf{z})\right).
\end{equation}
Here, JS divergence is defined as 
\begin{align}
&D_{JS_{\pi}^{M+1}}\left(\left\{q_{\phi_{m}}\left(\mathbf{z} \mid \mathbf{x}_{m}\right)\right\}_{m=1}^{M}, p(\mathbf{z})\right)\nonumber\\
&=\pi_{M+1} D_{KL}(p(\mathbf{z})||p_{g}(\mathbf{z}|X))+\sum_{m=1}^{M}\pi_m D_{KL}(q_{\phi_{m}}(\mathbf{z}|\mathbf{x}_m)||p_{g}(\mathbf{z}|X)),
\end{align}
where $\sum_{m=1}^{M+1}\pi_m=1$ and $p_{g}(\mathbf{z}|X)=g\left(\left\{q_{\phi_{m}}\left(\mathbf{z} \mid \mathbf{x}_{m}\right)\right\}_{m=1}^{M}, p(\mathbf{z})\right)$ is called the dynamic prior defined by the function $g$, which is set as the weighted PoE. 
Sutter et al.~\cite{sutter2020multimodal} show that mmJSD outperforms MVAE and MMVAE in assessing cross-modal generation and inference to latent representation.

The disadvantage of MoE is that aggregating experts does not result in a distribution that is sharper than the other experts. Therefore, even if we increase the number of experts, the shared representation does not become more informative as in PoE, which means that proper aggregated inference is not possible.

To mitigate the trade-off between PoE and MoE, Sutter et al.~\cite{sutter2021generalized} introduce a generalization of PoE and MoE, called mixture of products of experts (MoPoE) as follows.
\begin{equation}
q_{MoPoE}(\mathbf{z}|X) = \frac{1}{2^{M}} \sum_{X_k\in \mathcal{P}(X)}  q_{PoE}\left(\mathbf{z} \mid X_k\right),
\end{equation}
where $q_{PoE}\left(\mathbf{z} \mid X_k\right)\propto\prod_{\mathbf{x}_m \in X_k}q_{\phi_m}(\mathbf{z}|\mathbf{x}_m)$ and $\mathcal{P}(X)$ is a powerset of $M$ modalities. MoPoE has the advantages of both PoE and MoE (see Figure~\ref{fig:aggregated}(c)), and JVAE by this joint inference is called MoPoE-VAE.
However, because the number of PoE inferences increases exponentially with the number of modalities, the computational cost of ELBO for this model also increases exponentially.

Yadav et al.~\cite{yadav2020bridged} propose that another approach to approximating joint and unimodal inference from unimodal encoders is to introduce {\it bridge encoders} between latent variables of different modalities, such as associators in AVAE~\cite{jo2020associative}.
First, by letting $q_{\Phi}(\mathbf{z}|\mathbf{z}_1, \mathbf{z}_2)$ be the inference from latent representations from modalities, $\mathbf{z}_1=E_1(\mathbf{z}_1)$ and $\mathbf{z}_2=E_2(\mathbf{z}_2)$, to the shared representation $\mathbf{z}$, the joint inference is as follows:
\begin{equation}
q(\mathbf{z}|\mathbf{x}_1, \mathbf{x}_2) = q_{\Phi}(\mathbf{z}|E_1(\mathbf{x}_1), E_2(\mathbf{x}_2)).
\end{equation}
Next, introducing the bridge encoder $E_{12}$ from $\mathbf{z}_1$ to  $\mathbf{z}_2$, the unimodal inference becomes
\begin{equation}
q(\mathbf{z}|\mathbf{x}_1) = q_{\Phi}(\mathbf{z}|E_1(\mathbf{x}_1), E_{12}(E_1(\mathbf{x}_1)).
\end{equation}
They first train encoders and generative models with a JVAE objective, and then freeze them and learn bridge encoders with the unimodal VAE objective.
Although this model requires bridge encoders and two stages of training, it has the advantage of requiring fewer networks than approaches using surrogate unimodal inference. 

\subsection{Modality-specific latent variables}

In JVAE, each modality is assumed to be generated from a single shared latent variable. However, in reality, different modalities should have modality-specific factors in addition to the common factors. Based on this idea, various models with modality-specific latent variables have been proposed.

\begin{figure*}[tb]
 		\begin{center}
		\includegraphics[width=6cm]{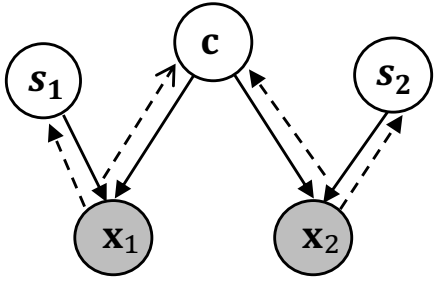}	
		\end{center}
			\caption{Graphical model of a multimodal generative model with modality-specific latent variables. In this figure, $\mathbf{c}$ is the modality-invariant latent variable, and $\mathbf{s}_1$ and $\mathbf{s}_2$ are the latent variables specific to $\mathbf{x}_1$ and $\mathbf{x}_2$, respectively. The factorization of the inference distribution represented by the dotted line differs depending on the model. MFM~\cite{tsai2018learning} has a label and additional latent variables called factors (see Equation~\eqref{eq:mfm})}
\label{fig:specific}
\end{figure*}

Basically, we decompose the shared latent variable $\mathbf{z}$ into the modality-specific latent variable $S=\{\mathbf{s}_m\}_{m=1}^M$ and the modality-independent latent variable $\mathbf{c}$, and assume the following generative model (see also Figure~\ref{fig:specific}):
\begin{eqnarray}
p_{\Theta}(X) =\int \int p_{\Theta}(X,S,\mathbf{c})dSd\mathbf{c} = \int  p(\mathbf{c})\int\prod_{m=1}^M p(\mathbf{s}_m)p_{\theta_m}(\mathbf{x}_m|\mathbf{s}_m,\mathbf{c})d\mathbf{s}_md\mathbf{c}.
\end{eqnarray}
A major challenge under this generative model is how to disentangle and infer modality-specific and modality-invariant representations from multimodal data.

The partitioned variational autoencoder (PVAE) proposed by Hsu et al.~\cite{hsu2018disentangling} assumes that a joint inference distribution can be factorized as
\begin{equation}
q_{\Phi}(S,\mathbf{c}|X)=q_{\Phi^c}(\mathbf{c}|X)\prod_{m=1}^M q_{\Phi^s_m}(\mathbf{s}_m|X),
\label{eq:pvae}
\end{equation}
and each of unimodal inference distributions is factorized as 
\begin{equation}
q_{\Phi}(\mathbf{s}_m,\mathbf{c}|\mathbf{x}_m)=q_{\phi^s_m}(\mathbf{s}_m|\mathbf{x}_m)q_{\phi^c_m}(\mathbf{c}|\mathbf{x}_m).
\end{equation}
The objective of PVAE consists of that of JVAE plus the KL divergence between joint inference and unimodal inference as in JMVAE (called multimodal–unimodal coherence).
Furthermore, hinge loss-based objective named cross-modality semantic contrastiveness is added to this objective such that modality-independent latent variables of different modalities from the same example are similar and those from different samples are dissimilar.

Sutter et al.~\cite{sutter2020multimodal} factorize the joint posterior, based on multi-level variational autoencoder~\cite{bouchacourt2018multi}, as follows:
\begin{equation}
q_{\Phi}(S,\mathbf{c}|X)=q_{\Phi^c}(\mathbf{c}|X)\prod_{m=1}^M q_{\phi^s_m}(\mathbf{s}_m|\mathbf{x}_m).
\end{equation}
The difference from joint inference in PVAE (Equation~\ref{eq:pvae}) is that each modality-specific latent variable is inferred from its modality input, not from all modalities. Moreover, they assume that $q_{\Phi^c}(\mathbf{c}|X)$ is approximated by aggregating unimodal inferences $\{q_{\phi^c_m}(\mathbf{c}|\mathbf{x}_m)\}_{m=1}^M$ in the form of PoE or MoE.

Lee et al.~\cite{lee2020private} approximates $q_{\Phi^c}(\mathbf{c}|X)$ by a PoE form and learns to optimize not only the objective of the JVAE, but also the objective that promotes cross-model generation and self-reconstruction by each PoE expert $q_{\phi^c_m}(\mathbf{c}|\mathbf{x}_m)$.
Moreover, that study applies the $\beta$-TCVAE decomposition to the KL divergence and penalizes the term of total correlation to disentangle modality-invariant and modality-specific latent variables.
To ensure that $\mathbf{c}$ is a discrete representation, concrete distribution~\cite{maddison2016concrete, jang2016categorical} is used as the inference distribution.

Daunhawer et al.~\cite{daunhawer2021self} introduce the following two objective in addition to the objective of JVAE.
The first objective is the mutual information between the multimodal input $X$ and the modality-invariant representation $\mathbf{c}$, which imposes the inclusion of multimodal information in $\mathbf{c}$.
However, since it is difficult to compute this mutual information directly, they estimate its lower bound using the sample-based InfoNCE estimator~\cite{oord2018representation}.
The second is the total correlation between modality-specific and modality-invariant latent variables, which is made independent of each other.
Unlike Lee et al.~\cite{lee2020private}, this total correlation is explicitly estimated by introducing and learning a discriminator based on the density-ratio trick~\cite{sugiyama2012density}.

Multimodal Factorization Model (MFM)~\cite{tsai2018learning} includes a label $\mathbf{y}$ as well as a multimodal input $X$ as observation variables, and assumes the following generative model:
\begin{align}
p_{\Theta}(X,\mathbf{y}) &=\int \int\int \int p_{\Theta}(X,\mathbf{y},S,\mathbf{c}, F^s, \mathbf{f}^c)dSd\mathbf{c}dF^sd\mathbf{f}^c\nonumber\\
&= \int \int  p(\mathbf{c})p_{\theta^{fc}}(\mathbf{y}|\mathbf{f}^c)p_{\theta^{c}}(\mathbf{f}^c|\mathbf{c})\int \int \prod_{m=1}^M p(\mathbf{s}_m)p_{\theta^{fs}_m}(\mathbf{x}_m|\mathbf{f}_m^s,\mathbf{f}^c)p_{\theta^{s}_m}(\mathbf{f}^s_{m}|\mathbf{s}_m)d\mathbf{f}_m^sd\mathbf{s}_md\mathbf{f}^cd\mathbf{c},
\label{eq:mfm}
\end{align}
where $F^s=\{\mathbf{f}^s_{m}\}_{m=1}^M$ and $\mathbf{f}^c$ are modality-specific and modality-invariant latent factors, respectively.
The inference distribution for MFM $q_{\Phi}(\mathbf{c}, S|X)$ is approximated in the same way as in the above studies.
The inference and generative models are optimized with the objective of Wasserstein autoencoders~\cite{tolstikhin2017wasserstein}.
In addition, Tsai et al.~\cite{tsai2018learning} introduce a surrogate inference distribution to handle missing modalities, such as JMVAE, and learn it so that cross-modal generation is optimal.

\subsection{Weakly-supervised learning}
In general, training a joint VAE requires that all modality combinations are always given, because joint inference is involved. However, it is difficult to prepare a large amount of multimodal data, and many examples in the training set obtained in the real environment might be sparse, i.e., some of the modalities are missing. Therefore, it is required to be able to train in weakly supervised settings.

Wu et al.~\cite{wu2018multimodal} show that the sub-sampling training scheme in MVAE allows a subset of the modalities to be used for training, i.e., weakly-supervised learning.

Shi et al.~\cite{shi2021relating} propose learning to learn whether different modalities are related and to maximize the following contrastive-style objective based on the max-margin metric.
\begin{equation}
\log p_{\Theta}(\mathbf{x}_1, \mathbf{x}_2) - \frac{1}{2}\left(\log \sum_{\mathbf{x}_1\in\{\mathbf{x}'_1\}_{i=1}^{N'}} p_{\Theta}(\mathbf{x}_1, \mathbf{x}_2)
+\log \sum_{\mathbf{x}_2\in\{\mathbf{x}'_2\}_{i=1}^{N'}}p_{\Theta}(\mathbf{x}_1, \mathbf{x}_2)
\right),
\end{equation}
where $\sum_{\mathbf{x}_m\in\{\mathbf{x}'_m\}_{i=1}^{N'}}$ is the sum of $N'$ negative samples of $\mathbf{x}_m$, and the joint log-likelihood in this objective is estimated by IWAE or $\chi $ upper bound~\cite{dieng2016variational}, rather than ELBO.
Shi et al.~\cite{shi2021relating} show that this method performs well not only with regular multimodal learning, but also given a small amount of paired multimodal data.

\section{Benchmark datasets and application}\label{sec:6}
Multimodal deep generative models have no multimodal benchmark that is always used like zero-shot learning. The most commonly used are image datasets such as MNIST~\cite{lecun1998gradient}, FashionMNIST~\cite{xiao2017/online}, and the street view house number (SVHN)~\cite{netzer2011reading}, which are often used as toy problems for deep learning.
However, since these datasets are not multimodal data, many studies turned them into multimodal settings in various ways. For example, the image and label are considered different modalities~\cite{suzuki2016joint, wu2018multimodal,wu2019multimodal, yadav2020bridged}, and the image with different noise is considered as multimodal data~\cite{wang2015deep, sutter2021generalized}. Another common practice is to pair multiple datasets that share the same label domain (i.e., number of classes) and consider them as multimodal data, such as MNIST and FashionMNIST~\cite{tian2019latent, jo2019cross, yadav2020bridged} or MNIST and SVHN~\cite{shi2019variational, daunhawer2021self, shi2021relating}. In addition, some studies have created multimodal data by adding information such as voice and text to these datasets~\cite{hsu2018disentangling, sutter2020multimodal}.

Several studies have experimented with a true multimodal setting rather than the toy problem described above.
A common multimodal dataset used in many multimodal deep generative models~\cite{suzuki2016joint, vedantam2017generative, sutter2020multimodal, sutter2020multimodal, yadav2020bridged, sutter2021generalized, hsu2018disentangling} is the CelebA dataset~\cite{liu2015faceattributes}, which contains face images and their attributes.
In CADA-VAE~\cite{schonfeld2019generalized}, the authors used the Caltech-UCSD Birds (CUB)-200 dataset~\cite{WelinderEtal2010}, which contains bird images and their attributes, as a benchmark for zero-shot learning. 
Shi et al.~\cite{shi2019variational,shi2021relating} used the CUB-200-2011 dataset~\cite{WahCUB_200_2011}, which is an extended version of the CUB-200 dataset and includes fine-grained natural language descriptions.
Jo et al.~\cite{jo2019cross} worked on hand pose estimation using RGB images and keypoints of each hand in the Rendered Hand pose dataset~\cite{zimmermann2017learning} as multimodal data.
Yin et al.~\cite{yin2017associate} trained their multimodal deep generative model using the handwriting motion and the corresponding handwriting image in the UJI Char Pen dataset as different modalities to confirm that cross-modal generation is possible. 
Tsai et al.~\cite{tsai2018learning} conducted experiments on real-world time-series multimodal datasets such as CMU-MOSI~\cite{zadeh2016multimodal}, which includes three modalities: language, acoustic, and visual.

Moreover, beyond the multimodal deep generative model defined in this study, i.e., in a setting where not all modalities are inferred and generated, various multimodal tasks are performed with VAEs, such as audio-visual speech enhancement~\cite{sadeghi2020robust, sadeghi2020audio} and the acquisition of the joint representation of depth and RGB images~\cite{bloesch2018codeslam}.

There have also been several studies on the application of multimodal deep generative models to robots.
Zambelli et al.~\cite{zambelli2020multimodal} trained a JVAE using five modalities from an iCub robot~\cite{metta2008icub} interacting with a piano keyboard as input, including joint positions and vision, to reconstruct missing modalities and predict their own sensorimotor states and others’ visual trajectories. 
Park et al.~\cite{park2018multimodal} introduced a LSTM-based JVAE to detect anomalous feeding executions given 17 sensory signals from five types of sensors on a PR2 robot.
Meo et al.~\cite{meo2021multimodal} proposed considering the joint angles in a simulated robot arm and the images provided by the camera as different modalities, and using JVAE to learn them and control the robot based on the active inference.
Korthals et al.~\cite{korthals2019multi} argued that the representation of the multimodal deep generative model can be used as a state as well as a reward in deep reinforcement learning, and showed that the proposed  model is effective in multi-agent reinforcement learning with multiple robots.

\section{Future challenges}\label{sec:7}
In this section we discuss future research directions for multimodal deep generative models.

The studies of multimodal deep generative models described so far have been validated mainly in two, or at most three, modalities.
However, the number of modalities in the real world is far greater, and more modality information has been used in previous studies of cognitive architecture based on probabilistic generative models.
As described in Section \ref{sec:related}, Nakamura et al.~\cite{nakamura2009grounding} used visual, tactile, auditory, and word information obtained from the robot.

Moreover, the greater the number and variety of modalities, the more difficult training the entire model end-to-end will be. Therefore, it is important to learn modules that deal with different modalities separately and integrate their inference. Coordinated models often use these two-step approaches; however, as mentioned earlier, they cannot perform inference on shared representations from arbitrary sets of modalities.
Recently, Symbol Emergence in
Robotics tool KIT (SERKET)~\cite{nakamura2018serket} and Neuro-SERKET~\cite{taniguchi2020neuro} have been proposed as a frameworks for integration in probabilistic models that deal with multimodal information. 
SERKET provides a protocol that divides inference from different modalities into the inference of modules of individual modalities and their communication, i.e., message-passing.
Neuro-SERKET, an extension of the SERKET framework for including neural networks, can integrate modules of different modalities that perform different inference procedures, such as Gibbs sampling and variational inference.
Another possible approach is to refer to the idea of global workspace (GW) theory in cognitive science~\cite{baars1993cognitive}, in which interactions between different modules are realized by a shared space that can be modified by any module and broadcast to all modules.
Goyal et al.~\cite{goyal2021coordination} proposed an attention-based GW architecture for communicating positions and modules in Transformer~\cite{vaswani2017attention} and slot-based modular architectures~\cite{goyal2019recurrent}. Such idea might also be applied to the integration of modules of different modalities.

Furthermore, multimodal deep generative models can be applied to the domain of world models~\cite{ha2018recurrent}, i.e., model-based reinforcement learning with self-supervised learning. Many studies on world models have used only images as input modalities~\cite{ha2018recurrent,hafner2019learning,hafner2019dream},  but we humans are building more reliable models in our brains of how the world is organized from many different types of information.
Recently, Taniguchi et al.~\cite{taniguchi2021whole} proposed implementing  prediction and decision making from multimodal information in the framework of generative models based on human cognitive systems.

From the viewpoint of robot research, as mentioned in Section 6, some studies used multimodal deep generative models, but they are not yet mainstream and their latest methods are rarely used. It is inevitable for robots to deal with multiple modalities and the ability of multimodal deep generative models to obtain integrated representations from them, to deal with missing modalities, and to transform between different modalities should help robots make better decisions in the real world.
However, there are several challenges in applying multimodal deep generative models to robots operating in the real world.
For example, in order for robots with multimodal deep generative models to be able to generalize properly in different environments, we need to acquire a large amount of training data for all environments, which is difficult to do in practice. To deal with this difficulty, it might be important to use techniques such as transfer learning and continuous learning~\cite{thrun1995lifelong,lesort2020continual}, given that humans have the ability to act in unknown environments based on their memories of other environments and the ability to continuously learn new things from the past.
Furthermore, in terms of model implementation, multimodal deep generative models might become larger and more complex due to the need to train large environments based on a large number of multimodal information, which makes it difficult to maintain and handle these implementations\footnote{Recall that deep generative models are generative models whose distributions are parameterized by deep neural networks; therefore, if these models are large, their implementation can be much more complex than the implementation of regular generative models.}.
Probabilistic modeling languages, such as Pyro~\cite{bingham2018pyro}, and deep generative modeling libraries, such as Pixyz~\cite{suzuki2021pixyz}, might help to address this difficulty.
We hope that multimodal deep generative models will be widely used in robot control in the future.

\section{Conclusion}\label{sec:8}
In this paper, we surveyed multimodal deep generative models in two categories: coordinated models and joint models. 
We classified the coordinated models into two criteria according to the how they define the closeness between inference distributions.
For the joint models, we summarized the study with three important challenges.
In particular, for the first challenge, the treatment of missing modalities, we described two approaches: the introduction of surrogate unimodal inference and the aggregation of unimodal inference.
In addition, we summarized the benchmark datasets and applications of multimodal deep generative models and discussed future directions.
We hope that this paper will be of help for future research on multimodal deep generative models.

\section*{Acknowledgements}
This paper is based on results obtained from a project, JPNP16007, subsidized by the New Energy and Industrial Technology Development Organization (NEDO).

\bibliographystyle{plain}
\bibliography{tADRguide.bib}

\begin{thebibliography}{100}

\bibitem{alayrac2020self}
Jean-Baptiste Alayrac, Adri{\`a} Recasens, Rosalia Schneider, Relja
  Arandjelovi{\'c}, Jason Ramapuram, Jeffrey De~Fauw, Lucas Smaira, Sander
  Dieleman, and Andrew Zisserman.
\newblock Self-supervised multimodal versatile networks.
\newblock {\em arXiv preprint arXiv:2006.16228}, 2020.

\bibitem{araki2012online}
Takaya Araki, Tomoaki Nakamura, Takayuki Nagai, Shogo Nagasaka, Tadahiro
  Taniguchi, and Naoto Iwahashi.
\newblock Online learning of concepts and words using multimodal lda and
  hierarchical pitman-yor language model.
\newblock In {\em 2012 IEEE/RSJ International Conference on Intelligent Robots
  and Systems}, pages 1623--1630. IEEE, 2012.

\bibitem{atrey2010multimodal}
Pradeep~K Atrey, M~Anwar Hossain, Abdulmotaleb El~Saddik, and Mohan~S
  Kankanhalli.
\newblock Multimodal fusion for multimedia analysis: a survey.
\newblock {\em Multimedia systems}, 16(6):345--379, 2010.

\bibitem{baars1993cognitive}
Bernard~J Baars.
\newblock {\em A cognitive theory of consciousness}.
\newblock Cambridge University Press, 1993.

\bibitem{baltruvsaitis2018multimodal}
Tadas Baltru{\v{s}}aitis, Chaitanya Ahuja, and Louis-Philippe Morency.
\newblock Multimodal machine learning: A survey and taxonomy.
\newblock {\em IEEE transactions on pattern analysis and machine intelligence},
  41(2):423--443, 2018.

\bibitem{bengio2013representation}
Yoshua Bengio, Aaron Courville, and Pascal Vincent.
\newblock Representation learning: A review and new perspectives.
\newblock {\em IEEE transactions on pattern analysis and machine intelligence},
  35(8):1798--1828, 2013.

\bibitem{bingham2018pyro}
Eli Bingham, Jonathan~P Chen, Martin Jankowiak, Fritz Obermeyer, Neeraj
  Pradhan, Theofanis Karaletsos, Rohit Singh, Paul Szerlip, Paul Horsfall, and
  Noah~D Goodman.
\newblock Pyro: Deep universal probabilistic programming.
\newblock {\em arXiv preprint arXiv:1810.09538}, 2018.

\bibitem{blei2003modeling}
David~M Blei and Michael~I Jordan.
\newblock Modeling annotated data.
\newblock In {\em Proceedings of the 26th annual international ACM SIGIR
  conference on Research and development in informaion retrieval}, pages
  127--134, 2003.

\bibitem{blei2003latent}
David~M Blei, Andrew~Y Ng, and Michael~I Jordan.
\newblock Latent dirichlet allocation.
\newblock {\em the Journal of machine Learning research}, 3:993--1022, 2003.

\bibitem{bliss1993synaptic}
Tim~VP Bliss and Graham~L Collingridge.
\newblock A synaptic model of memory: long-term potentiation in the
  hippocampus.
\newblock {\em Nature}, 361(6407):31--39, 1993.

\bibitem{bloesch2018codeslam}
Michael Bloesch, Jan Czarnowski, Ronald Clark, Stefan Leutenegger, and Andrew~J
  Davison.
\newblock Codeslam―learning a compact, optimisable representation for dense
  visual slam.
\newblock In {\em Proceedings of the IEEE conference on computer vision and
  pattern recognition}, pages 2560--2568, 2018.

\bibitem{bonneel2015sliced}
Nicolas Bonneel, Julien Rabin, Gabriel Peyr{\'e}, and Hanspeter Pfister.
\newblock Sliced and radon wasserstein barycenters of measures.
\newblock {\em Journal of Mathematical Imaging and Vision}, 51(1):22--45, 2015.

\bibitem{bouchacourt2018multi}
Diane Bouchacourt, Ryota Tomioka, and Sebastian Nowozin.
\newblock Multi-level variational autoencoder: Learning disentangled
  representations from grouped observations.
\newblock In {\em Proceedings of the AAAI Conference on Artificial
  Intelligence}, volume~32, 2018.

\bibitem{burda2015importance}
Yuri Burda, Roger Grosse, and Ruslan Salakhutdinov.
\newblock Importance weighted autoencoders.
\newblock {\em arXiv preprint arXiv:1509.00519}, 2015.

\bibitem{daunhawer2021self}
Imant Daunhawer, Thomas~M Sutter, Ri{\v{c}}ards Marcinkevi{\v{c}}s, and Julia~E
  Vogt.
\newblock Self-supervised disentanglement of modality-specific and shared
  factors improves multimodal generative models.
\newblock {\em Pattern Recognition}, 12544:459, 2021.

\bibitem{dieng2016variational}
Adji~B Dieng, Dustin Tran, Rajesh Ranganath, John Paisley, and David~M Blei.
\newblock Variational inference via $\chi $-upper bound minimization.
\newblock {\em arXiv preprint arXiv:1611.00328}, 2016.

\bibitem{gao2020survey}
Jing Gao, Peng Li, Zhikui Chen, and Jianing Zhang.
\newblock A survey on deep learning for multimodal data fusion.
\newblock {\em Neural Computation}, 32(5):829--864, 2020.

\bibitem{gershman2014amortized}
Samuel Gershman and Noah Goodman.
\newblock Amortized inference in probabilistic reasoning.
\newblock In {\em Proceedings of the annual meeting of the cognitive science
  society}, volume~36, 2014.

\bibitem{goodfellow2014generative}
Ian~J Goodfellow, Jean Pouget-Abadie, Mehdi Mirza, Bing Xu, David Warde-Farley,
  Sherjil Ozair, Aaron Courville, and Yoshua Bengio.
\newblock Generative adversarial networks.
\newblock {\em arXiv preprint arXiv:1406.2661}, 2014.

\bibitem{goyal2021coordination}
Anirudh Goyal, Aniket Didolkar, Alex Lamb, Kartikeya Badola, Nan~Rosemary Ke,
  Nasim Rahaman, Jonathan Binas, Charles Blundell, Michael Mozer, and Yoshua
  Bengio.
\newblock Coordination among neural modules through a shared global workspace.
\newblock {\em arXiv preprint arXiv:2103.01197}, 2021.

\bibitem{goyal2019recurrent}
Anirudh Goyal, Alex Lamb, Jordan Hoffmann, Shagun Sodhani, Sergey Levine,
  Yoshua Bengio, and Bernhard Sch{\"o}lkopf.
\newblock Recurrent independent mechanisms.
\newblock {\em arXiv preprint arXiv:1909.10893}, 2019.

\bibitem{ha2018recurrent}
David Ha and J{\"u}rgen Schmidhuber.
\newblock Recurrent world models facilitate policy evolution.
\newblock {\em arXiv preprint arXiv:1809.01999}, 2018.

\bibitem{hafner2019dream}
Danijar Hafner, Timothy Lillicrap, Jimmy Ba, and Mohammad Norouzi.
\newblock Dream to control: Learning behaviors by latent imagination.
\newblock {\em arXiv preprint arXiv:1912.01603}, 2019.

\bibitem{hafner2019learning}
Danijar Hafner, Timothy Lillicrap, Ian Fischer, Ruben Villegas, David Ha,
  Honglak Lee, and James Davidson.
\newblock Learning latent dynamics for planning from pixels.
\newblock In {\em International Conference on Machine Learning}, pages
  2555--2565. PMLR, 2019.

\bibitem{higgins2016beta}
Irina Higgins, Loic Matthey, Arka Pal, Christopher Burgess, Xavier Glorot,
  Matthew Botvinick, Shakir Mohamed, and Alexander Lerchner.
\newblock beta-vae: Learning basic visual concepts with a constrained
  variational framework.
\newblock 2016.

\bibitem{higgins2017scan}
Irina Higgins, Nicolas Sonnerat, Loic Matthey, Arka Pal, Christopher~P Burgess,
  Matko Bosnjak, Murray Shanahan, Matthew Botvinick, Demis Hassabis, and
  Alexander Lerchner.
\newblock Scan: Learning hierarchical compositional visual concepts.
\newblock {\em arXiv preprint arXiv:1707.03389}, 2017.

\bibitem{hinton2002training}
Geoffrey~E Hinton.
\newblock Training products of experts by minimizing contrastive divergence.
\newblock {\em Neural computation}, 14(8):1771--1800, 2002.

\bibitem{hinton2009deep}
Geoffrey~E Hinton.
\newblock Deep belief networks.
\newblock {\em Scholarpedia}, 4(5):5947, 2009.

\bibitem{hinton2006fast}
Geoffrey~E Hinton, Simon Osindero, and Yee-Whye Teh.
\newblock A fast learning algorithm for deep belief nets.
\newblock {\em Neural computation}, 18(7):1527--1554, 2006.

\bibitem{hinton2006reducing}
Geoffrey~E Hinton and Ruslan~R Salakhutdinov.
\newblock Reducing the dimensionality of data with neural networks.
\newblock {\em science}, 313(5786):504--507, 2006.

\bibitem{holyoak1987parallel}
Keith~J Holyoak.
\newblock Parallel distributed processing: explorations in the microstructure
  of cognition.
\newblock {\em Science}, 236:992--997, 1987.

\bibitem{hotelling1992relations}
Harold Hotelling.
\newblock Relations between two sets of variates.
\newblock In {\em Breakthroughs in statistics}, pages 162--190. Springer, 1992.

\bibitem{hsu2018disentangling}
Wei-Ning Hsu and James Glass.
\newblock Disentangling by partitioning: A representation learning framework
  for multimodal sensory data.
\newblock {\em arXiv preprint arXiv:1805.11264}, 2018.

\bibitem{huang2013audio}
Jing Huang and Brian Kingsbury.
\newblock Audio-visual deep learning for noise robust speech recognition.
\newblock In {\em 2013 IEEE International Conference on Acoustics, Speech and
  Signal Processing}, pages 7596--7599. IEEE, 2013.

\bibitem{isola2017image}
Phillip Isola, Jun-Yan Zhu, Tinghui Zhou, and Alexei~A Efros.
\newblock Image-to-image translation with conditional adversarial networks.
\newblock In {\em Proceedings of the IEEE conference on computer vision and
  pattern recognition}, pages 1125--1134, 2017.

\bibitem{ivanovic2020multimodal}
Boris Ivanovic, Karen Leung, Edward Schmerling, and Marco Pavone.
\newblock Multimodal deep generative models for trajectory prediction: A
  conditional variational autoencoder approach.
\newblock {\em IEEE Robotics and Automation Letters}, 6(2):295--302, 2020.

\bibitem{jang2016categorical}
Eric Jang, Shixiang Gu, and Ben Poole.
\newblock Categorical reparameterization with gumbel-softmax.
\newblock {\em arXiv preprint arXiv:1611.01144}, 2016.

\bibitem{jaques2017multimodal}
Natasha Jaques, Sara Taylor, Akane Sano, and Rosalind Picard.
\newblock Multimodal autoencoder: A deep learning approach to filling in
  missing sensor data and enabling better mood prediction.
\newblock In {\em 2017 Seventh International Conference on Affective Computing
  and Intelligent Interaction (ACII)}, pages 202--208. IEEE, 2017.

\bibitem{jo2019cross}
Dae~Ung Jo, ByeongJu Lee, Jongwon Choi, Haanju Yoo, and Jin~Young Choi.
\newblock Cross-modal variational auto-encoder with distributed latent spaces
  and associators.
\newblock {\em arXiv preprint arXiv:1905.12867}, 2019.

\bibitem{jo2020associative}
Dae~Ung Jo, ByeongJu Lee, Jongwon Choi, Haanju Yoo, and Jin~Young Choi.
\newblock Associative variational auto-encoder with distributed latent spaces
  and associators.
\newblock In {\em Proceedings of the AAAI Conference on Artificial
  Intelligence}, volume~34, pages 11197--11204, 2020.

\bibitem{kim2013deep}
Yelin Kim, Honglak Lee, and Emily~Mower Provost.
\newblock Deep learning for robust feature generation in audiovisual emotion
  recognition.
\newblock In {\em 2013 IEEE international conference on acoustics, speech and
  signal processing}, pages 3687--3691. IEEE, 2013.

\bibitem{kingma2018glow}
Diederik~P Kingma and Prafulla Dhariwal.
\newblock Glow: Generative flow with invertible 1x1 convolutions.
\newblock {\em arXiv preprint arXiv:1807.03039}, 2018.

\bibitem{kingma2013auto}
Diederik~P Kingma and Max Welling.
\newblock Auto-encoding variational bayes.
\newblock {\em arXiv preprint arXiv:1312.6114}, 2013.

\bibitem{korthals2019multi}
Timo Korthals, Daniel Rudolph, J{\"u}rgen Leitner, Marc Hesse, and Ulrich
  R{\"u}ckert.
\newblock Multi-modal generative models for learning epistemic active sensing.
\newblock In {\em 2019 International Conference on Robotics and Automation
  (ICRA)}, pages 3319--3325. IEEE, 2019.

\bibitem{lahat2015multimodal}
Dana Lahat, T{\"u}lay Adali, and Christian Jutten.
\newblock Multimodal data fusion: an overview of methods, challenges, and
  prospects.
\newblock {\em Proceedings of the IEEE}, 103(9):1449--1477, 2015.

\bibitem{lecun1998gradient}
Yann LeCun, L{\'e}on Bottou, Yoshua Bengio, and Patrick Haffner.
\newblock Gradient-based learning applied to document recognition.
\newblock {\em Proceedings of the IEEE}, 86(11):2278--2324, 1998.

\bibitem{lee2018diverse}
Hsin-Ying Lee, Hung-Yu Tseng, Jia-Bin Huang, Maneesh Singh, and Ming-Hsuan
  Yang.
\newblock Diverse image-to-image translation via disentangled representations.
\newblock In {\em Proceedings of the European conference on computer vision
  (ECCV)}, pages 35--51, 2018.

\bibitem{lee2020private}
Mihee Lee and Vladimir Pavlovic.
\newblock Private-shared disentangled multimodal vae for learning of hybrid
  latent representations.
\newblock {\em arXiv preprint arXiv:2012.13024}, 2020.

\bibitem{lesort2020continual}
Timoth{\'e}e Lesort, Vincenzo Lomonaco, Andrei Stoian, Davide Maltoni, David
  Filliat, and Natalia D{\'\i}az-Rodr{\'\i}guez.
\newblock Continual learning for robotics: Definition, framework, learning
  strategies, opportunities and challenges.
\newblock {\em Information fusion}, 58:52--68, 2020.

\bibitem{li2018survey}
Yingming Li, Ming Yang, and Zhongfei Zhang.
\newblock A survey of multi-view representation learning.
\newblock {\em IEEE transactions on knowledge and data engineering},
  31(10):1863--1883, 2018.

\bibitem{liu2017unsupervised}
Ming-Yu Liu, Thomas Breuel, and Jan Kautz.
\newblock Unsupervised image-to-image translation networks.
\newblock {\em arXiv preprint arXiv:1703.00848}, 2017.

\bibitem{liu2015faceattributes}
Ziwei Liu, Ping Luo, Xiaogang Wang, and Xiaoou Tang.
\newblock Deep learning face attributes in the wild.
\newblock In {\em Proceedings of International Conference on Computer Vision
  (ICCV)}, December 2015.

\bibitem{maddison2016concrete}
Chris~J Maddison, Andriy Mnih, and Yee~Whye Teh.
\newblock The concrete distribution: A continuous relaxation of discrete random
  variables.
\newblock {\em arXiv preprint arXiv:1611.00712}, 2016.

\bibitem{meo2021multimodal}
Cristian Meo and Pablo Lanillos.
\newblock Multimodal vae active inference controller.
\newblock {\em arXiv preprint arXiv:2103.04412}, 2021.

\bibitem{metta2008icub}
Giorgio Metta, Giulio Sandini, David Vernon, Lorenzo Natale, and Francesco
  Nori.
\newblock The icub humanoid robot: an open platform for research in embodied
  cognition.
\newblock In {\em Proceedings of the 8th workshop on performance metrics for
  intelligent systems}, pages 50--56, 2008.

\bibitem{mirza2014conditional}
Mehdi Mirza and Simon Osindero.
\newblock Conditional generative adversarial nets.
\newblock {\em arXiv preprint arXiv:1411.1784}, 2014.

\bibitem{nakamura2014mutual}
Tomoaki Nakamura, Takayuki Nagai, Kotaro Funakoshi, Shogo Nagasaka, Tadahiro
  Taniguchi, and Naoto Iwahashi.
\newblock Mutual learning of an object concept and language model based on mlda
  and npylm.
\newblock In {\em 2014 IEEE/RSJ International Conference on Intelligent Robots
  and Systems}, pages 600--607. IEEE, 2014.

\bibitem{nakamura2009grounding}
Tomoaki Nakamura, Takayuki Nagai, and Naoto Iwahashi.
\newblock Grounding of word meanings in multimodal concepts using lda.
\newblock In {\em 2009 IEEE/RSJ International Conference on Intelligent Robots
  and Systems}, pages 3943--3948. IEEE, 2009.

\bibitem{nakamura2011bag}
Tomoaki Nakamura, Takayuki Nagai, and Naoto Iwahashi.
\newblock Bag of multimodal lda models for concept formation.
\newblock In {\em 2011 IEEE International Conference on Robotics and
  Automation}, pages 6233--6238. IEEE, 2011.

\bibitem{nakamura2018serket}
Tomoaki Nakamura, Takayuki Nagai, and Tadahiro Taniguchi.
\newblock Serket: An architecture for connecting stochastic models to realize a
  large-scale cognitive model.
\newblock {\em Frontiers in neurorobotics}, 12:25, 2018.

\bibitem{netzer2011reading}
Yuval Netzer, Tao Wang, Adam Coates, Alessandro Bissacco, Bo~Wu, and Andrew~Y
  Ng.
\newblock Reading digits in natural images with unsupervised feature learning.
\newblock 2011.

\bibitem{ngiam2011multimodal}
Jiquan Ngiam, Aditya Khosla, Mingyu Kim, Juhan Nam, Honglak Lee, and Andrew~Y
  Ng.
\newblock Multimodal deep learning.
\newblock In {\em ICML}, 2011.

\bibitem{noda2014multimodal}
Kuniaki Noda, Hiroaki Arie, Yuki Suga, and Tetsuya Ogata.
\newblock Multimodal integration learning of robot behavior using deep neural
  networks.
\newblock {\em Robotics and Autonomous Systems}, 62(6):721--736, 2014.

\bibitem{oord2016wavenet}
Aaron van~den Oord, Sander Dieleman, Heiga Zen, Karen Simonyan, Oriol Vinyals,
  Alex Graves, Nal Kalchbrenner, Andrew Senior, and Koray Kavukcuoglu.
\newblock Wavenet: A generative model for raw audio.
\newblock {\em arXiv preprint arXiv:1609.03499}, 2016.

\bibitem{oord2018representation}
Aaron van~den Oord, Yazhe Li, and Oriol Vinyals.
\newblock Representation learning with contrastive predictive coding.
\newblock {\em arXiv preprint arXiv:1807.03748}, 2018.

\bibitem{ouyang2014multi}
Wanli Ouyang, Xiao Chu, and Xiaogang Wang.
\newblock Multi-source deep learning for human pose estimation.
\newblock In {\em Proceedings of the IEEE Conference on Computer Vision and
  Pattern Recognition}, pages 2329--2336, 2014.

\bibitem{pan2009survey}
Sinno~Jialin Pan and Qiang Yang.
\newblock A survey on transfer learning.
\newblock {\em IEEE Transactions on knowledge and data engineering},
  22(10):1345--1359, 2009.

\bibitem{pang2015deep}
Lei Pang, Shiai Zhu, and Chong-Wah Ngo.
\newblock Deep multimodal learning for affective analysis and retrieval.
\newblock {\em IEEE Transactions on Multimedia}, 17(11):2008--2020, 2015.

\bibitem{park2018multimodal}
Daehyung Park, Yuuna Hoshi, and Charles~C Kemp.
\newblock A multimodal anomaly detector for robot-assisted feeding using an
  lstm-based variational autoencoder.
\newblock {\em IEEE Robotics and Automation Letters}, 3(3):1544--1551, 2018.

\bibitem{pourpanah2020review}
Farhad Pourpanah, Moloud Abdar, Yuxuan Luo, Xinlei Zhou, Ran Wang, Chee~Peng
  Lim, and Xi-Zhao Wang.
\newblock A review of generalized zero-shot learning methods.
\newblock {\em arXiv preprint arXiv:2011.08641}, 2020.

\bibitem{putthividhy2010topic}
Duangmanee Putthividhy, Hagai~T Attias, and Srikantan~S Nagarajan.
\newblock Topic regression multi-modal latent dirichlet allocation for image
  annotation.
\newblock In {\em 2010 IEEE Computer Society Conference on Computer Vision and
  Pattern Recognition}, pages 3408--3415. IEEE, 2010.

\bibitem{rezende2015variational}
Danilo Rezende and Shakir Mohamed.
\newblock Variational inference with normalizing flows.
\newblock In {\em International Conference on Machine Learning}, pages
  1530--1538. PMLR, 2015.

\bibitem{sadeghi2020robust}
Mostafa Sadeghi and Xavier Alameda-Pineda.
\newblock Robust unsupervised audio-visual speech enhancement using a mixture
  of variational autoencoders.
\newblock In {\em ICASSP 2020-2020 IEEE International Conference on Acoustics,
  Speech and Signal Processing (ICASSP)}, pages 7534--7538. IEEE, 2020.

\bibitem{sadeghi2020audio}
Mostafa Sadeghi, Simon Leglaive, Xavier Alameda-Pineda, Laurent Girin, and Radu
  Horaud.
\newblock Audio-visual speech enhancement using conditional variational
  auto-encoders.
\newblock {\em IEEE/ACM Transactions on Audio, Speech, and Language
  Processing}, 28:1788--1800, 2020.

\bibitem{salakhutdinov2009deep}
Ruslan Salakhutdinov and Geoffrey Hinton.
\newblock Deep boltzmann machines.
\newblock In {\em Artificial intelligence and statistics}, pages 448--455.
  PMLR, 2009.

\bibitem{schonfeld2019generalized}
Edgar Schonfeld, Sayna Ebrahimi, Samarth Sinha, Trevor Darrell, and Zeynep
  Akata.
\newblock Generalized zero-and few-shot learning via aligned variational
  autoencoders.
\newblock In {\em Proceedings of the IEEE/CVF Conference on Computer Vision and
  Pattern Recognition}, pages 8247--8255, 2019.

\bibitem{shi2010transfer}
Xiaoxiao Shi, Qi~Liu, Wei Fan, S~Yu Philip, and Ruixin Zhu.
\newblock Transfer learning on heterogenous feature spaces via spectral
  transformation.
\newblock In {\em 2010 IEEE international conference on data mining}, pages
  1049--1054. IEEE, 2010.

\bibitem{shi2021relating}
Yuge Shi, Brooks Paige, Philip Torr, and Siddharth N.
\newblock Relating by contrasting: A data-efficient framework for multimodal
  generative models.
\newblock In {\em International Conference on Learning Representations}, 2021.

\bibitem{shi2019variational}
Yuge Shi, Narayanaswamy Siddharth, Brooks Paige, and Philip Torr.
\newblock Variational mixture-of-experts autoencoders for multi-modal deep
  generative models.
\newblock In {\em Advances in Neural Information Processing Systems}, pages
  15718--15729, 2019.

\bibitem{sohn2015learning}
Kihyuk Sohn, Honglak Lee, and Xinchen Yan.
\newblock Learning structured output representation using deep conditional
  generative models.
\newblock {\em Advances in neural information processing systems},
  28:3483--3491, 2015.

\bibitem{sohn2014improved}
Kihyuk Sohn, Wenling Shang, and Honglak Lee.
\newblock Improved multimodal deep learning with variation of information.
\newblock {\em Advances in neural information processing systems},
  27:2141--2149, 2014.

\bibitem{srivastava2012learning}
Nitish Srivastava and Ruslan Salakhutdinov.
\newblock Learning representations for multimodal data with deep belief nets.
\newblock In {\em International conference on machine learning workshop},
  volume~79, page~3, 2012.

\bibitem{srivastava2012multimodal}
Nitish Srivastava, Ruslan Salakhutdinov, et~al.
\newblock Multimodal learning with deep boltzmann machines.
\newblock In {\em NIPS}, volume~1, page~2. Citeseer, 2012.

\bibitem{stein1993merging}
Barry~E Stein and M~Alex Meredith.
\newblock {\em The merging of the senses.}
\newblock 1993.

\bibitem{sugiyama2012density}
Masashi Sugiyama, Taiji Suzuki, and Takafumi Kanamori.
\newblock Density-ratio matching under the bregman divergence: a unified
  framework of density-ratio estimation.
\newblock {\em Annals of the Institute of Statistical Mathematics},
  64(5):1009--1044, 2012.

\bibitem{suk2014hierarchical}
Heung-Il Suk, Seong-Whan Lee, Dinggang Shen, Alzheimer's Disease~Neuroimaging
  Initiative, et~al.
\newblock Hierarchical feature representation and multimodal fusion with deep
  learning for ad/mci diagnosis.
\newblock {\em NeuroImage}, 101:569--582, 2014.

\bibitem{sun2013survey}
Shiliang Sun.
\newblock A survey of multi-view machine learning.
\newblock {\em Neural computing and applications}, 23(7):2031--2038, 2013.

\bibitem{sutter2020multimodal}
Thomas~M Sutter, Imant Daunhawer, and Julia~E Vogt.
\newblock Multimodal generative learning utilizing jensen-shannon-divergence.
\newblock {\em arXiv preprint arXiv:2006.08242}, 2020.

\bibitem{sutter2021generalized}
Thomas~M. Sutter, Imant Daunhawer, and Julia~E Vogt.
\newblock Generalized multimodal {ELBO}.
\newblock In {\em International Conference on Learning Representations}, 2021.

\bibitem{suzuki2021pixyz}
Masahiro Suzuki, Takaaki Kaneko, and Yutaka Matsuo.
\newblock Pixyz: a library for developing deep generative models, 2021.

\bibitem{suzuki2016joint}
Masahiro Suzuki, Kotaro Nakayama, and Yutaka Matsuo.
\newblock Joint multimodal learning with deep generative models.
\newblock {\em arXiv preprint arXiv:1611.01891}, 2016.

\bibitem{suzuki2018improving}
Masahiro Suzuki, Kotaro Nakayama, and Yutaka Matsuo.
\newblock Improving bi-directional generation between different modalities with
  variational autoencoders.
\newblock {\em arXiv preprint arXiv:1801.08702}, 2018.

\bibitem{taniguchi2020neuro}
Tadahiro Taniguchi, Tomoaki Nakamura, Masahiro Suzuki, Ryo Kuniyasu, Kaede
  Hayashi, Akira Taniguchi, Takato Horii, and Takayuki Nagai.
\newblock Neuro-serket: development of integrative cognitive system through the
  composition of deep probabilistic generative models.
\newblock {\em New Generation Computing}, pages 1--26, 2020.

\bibitem{taniguchi2021whole}
Tadahiro Taniguchi, Hiroshi Yamakawa, Takayuki Nagai, Kenji Doya, Masamichi
  Sakagami, Masahiro Suzuki, Tomoaki Nakamura, and Akira Taniguchi.
\newblock Whole brain probabilistic generative model toward realizing cognitive
  architecture for developmental robots.
\newblock {\em arXiv preprint arXiv:2103.08183}, 2021.

\bibitem{thrun1995lifelong}
Sebastian Thrun and Tom~M Mitchell.
\newblock Lifelong robot learning.
\newblock {\em Robotics and autonomous systems}, 15(1-2):25--46, 1995.

\bibitem{tian2019latent}
Yingtao Tian and Jesse Engel.
\newblock Latent translation: Crossing modalities by bridging generative
  models.
\newblock {\em arXiv preprint arXiv:1902.08261}, 2019.

\bibitem{tian2019contrastive}
Yonglong Tian, Dilip Krishnan, and Phillip Isola.
\newblock Contrastive multiview coding.
\newblock {\em arXiv preprint arXiv:1906.05849}, 2019.

\bibitem{tolstikhin2017wasserstein}
Ilya Tolstikhin, Olivier Bousquet, Sylvain Gelly, and Bernhard Schoelkopf.
\newblock Wasserstein auto-encoders.
\newblock {\em arXiv preprint arXiv:1711.01558}, 2017.

\bibitem{tsai2018learning}
Yao-Hung~Hubert Tsai, Paul~Pu Liang, Amir Zadeh, Louis-Philippe Morency, and
  Ruslan Salakhutdinov.
\newblock Learning factorized multimodal representations.
\newblock {\em arXiv preprint arXiv:1806.06176}, 2018.

\bibitem{tsai2021selfsupervised}
Yao-Hung~Hubert Tsai, Yue Wu, Ruslan Salakhutdinov, and Louis-Philippe Morency.
\newblock Self-supervised learning from a multi-view perspective.
\newblock In {\em International Conference on Learning Representations}, 2021.

\bibitem{tschannen2018recent}
Michael Tschannen, Olivier Bachem, and Mario Lucic.
\newblock Recent advances in autoencoder-based representation learning.
\newblock {\em arXiv preprint arXiv:1812.05069}, 2018.

\bibitem{van2016pixel}
Aaron Van~Oord, Nal Kalchbrenner, and Koray Kavukcuoglu.
\newblock Pixel recurrent neural networks.
\newblock In {\em International Conference on Machine Learning}, pages
  1747--1756. PMLR, 2016.

\bibitem{vaswani2017attention}
Ashish Vaswani, Noam Shazeer, Niki Parmar, Jakob Uszkoreit, Llion Jones,
  Aidan~N Gomez, Lukasz Kaiser, and Illia Polosukhin.
\newblock Attention is all you need.
\newblock {\em arXiv preprint arXiv:1706.03762}, 2017.

\bibitem{vedantam2017generative}
Ramakrishna Vedantam, Ian Fischer, Jonathan Huang, and Kevin Murphy.
\newblock Generative models of visually grounded imagination.
\newblock {\em arXiv preprint arXiv:1705.10762}, 2017.

\bibitem{vincent2010stacked}
Pascal Vincent, Hugo Larochelle, Isabelle Lajoie, Yoshua Bengio, Pierre-Antoine
  Manzagol, and L{\'e}on Bottou.
\newblock Stacked denoising autoencoders: Learning useful representations in a
  deep network with a local denoising criterion.
\newblock {\em Journal of machine learning research}, 11(12), 2010.

\bibitem{WahCUB_200_2011}
C.~Wah, S.~Branson, P.~Welinder, P.~Perona, and S.~Belongie.
\newblock {The Caltech-UCSD Birds-200-2011 Dataset}.
\newblock Technical report, 2011.

\bibitem{wang2015deep}
Daixin Wang, Peng Cui, Mingdong Ou, and Wenwu Zhu.
\newblock Deep multimodal hashing with orthogonal regularization.
\newblock In {\em Twenty-Fourth International Joint Conference on Artificial
  Intelligence}, 2015.

\bibitem{wang2016comprehensive}
Kaiye Wang, Qiyue Yin, Wei Wang, Shu Wu, and Liang Wang.
\newblock A comprehensive survey on cross-modal retrieval.
\newblock {\em arXiv preprint arXiv:1607.06215}, 2016.

\bibitem{wang2016deep}
Weiran Wang, Xinchen Yan, Honglak Lee, and Karen Livescu.
\newblock Deep variational canonical correlation analysis.
\newblock {\em arXiv preprint arXiv:1610.03454}, 2016.

\bibitem{weiss2016survey}
Karl Weiss, Taghi~M Khoshgoftaar, and DingDing Wang.
\newblock A survey of transfer learning.
\newblock {\em Journal of Big data}, 3(1):9, 2016.

\bibitem{WelinderEtal2010}
P.~Welinder, S.~Branson, T.~Mita, C.~Wah, F.~Schroff, S.~Belongie, and
  P.~Perona.
\newblock {Caltech-UCSD Birds 200}.
\newblock Technical Report CNS-TR-2010-001, California Institute of Technology,
  2010.

\bibitem{wu2018multimodal}
Mike Wu and Noah Goodman.
\newblock Multimodal generative models for scalable weakly-supervised learning.
\newblock In {\em Advances in Neural Information Processing Systems}, pages
  5575--5585, 2018.

\bibitem{wu2019multimodal}
Mike Wu and Noah Goodman.
\newblock Multimodal generative models for compositional representation
  learning.
\newblock {\em arXiv preprint arXiv:1912.05075}, 2019.

\bibitem{xian2017zero}
Yongqin Xian, Bernt Schiele, and Zeynep Akata.
\newblock Zero-shot learning-the good, the bad and the ugly.
\newblock In {\em Proceedings of the IEEE Conference on Computer Vision and
  Pattern Recognition}, pages 4582--4591, 2017.

\bibitem{xiao2017/online}
Han Xiao, Kashif Rasul, and Roland Vollgraf.
\newblock Fashion-mnist: a novel image dataset for benchmarking machine
  learning algorithms, 2017.

\bibitem{xing2012mining}
Eric~P Xing, Rong Yan, and Alexander~G Hauptmann.
\newblock Mining associated text and images with dual-wing harmoniums.
\newblock {\em arXiv preprint arXiv:1207.1423}, 2012.

\bibitem{xu2013survey}
Chang Xu, Dacheng Tao, and Chao Xu.
\newblock A survey on multi-view learning.
\newblock {\em arXiv preprint arXiv:1304.5634}, 2013.

\bibitem{yadav2020bridged}
Ravindra Yadav, Ashish Sardana, Vinay Namboodiri, and Rajesh~M Hegde.
\newblock Bridged variational autoencoders for joint modeling of images and
  attributes.
\newblock In {\em Proceedings of the IEEE/CVF Winter Conference on Applications
  of Computer Vision}, pages 1479--1487, 2020.

\bibitem{yin2017associate}
Hang Yin, Francisco Melo, Aude Billard, and Ana Paiva.
\newblock Associate latent encodings in learning from demonstrations.
\newblock In {\em Proceedings of the AAAI Conference on Artificial
  Intelligence}, volume~31, 2017.

\bibitem{zadeh2016multimodal}
Amir Zadeh, Rowan Zellers, Eli Pincus, and Louis-Philippe Morency.
\newblock Multimodal sentiment intensity analysis in videos: Facial gestures
  and verbal messages.
\newblock {\em IEEE Intelligent Systems}, 31(6):82--88, 2016.

\bibitem{zambelli2020multimodal}
Martina Zambelli, Antoine Cully, and Yiannis Demiris.
\newblock Multimodal representation models for prediction and control from
  partial information.
\newblock {\em Robotics and Autonomous Systems}, 123:103312, 2020.

\bibitem{zhang2020multimodal}
Chao Zhang, Zichao Yang, Xiaodong He, and Li~Deng.
\newblock Multimodal intelligence: Representation learning, information fusion,
  and applications.
\newblock {\em IEEE Journal of Selected Topics in Signal Processing},
  14(3):478--493, 2020.

\bibitem{zhang2017stackgan}
Han Zhang, Tao Xu, Hongsheng Li, Shaoting Zhang, Xiaogang Wang, Xiaolei Huang,
  and Dimitris~N Metaxas.
\newblock Stackgan: Text to photo-realistic image synthesis with stacked
  generative adversarial networks.
\newblock In {\em Proceedings of the IEEE international conference on computer
  vision}, pages 5907--5915, 2017.

\bibitem{zheng2014topic}
Yin Zheng, Yu-Jin Zhang, and Hugo Larochelle.
\newblock Topic modeling of multimodal data: an autoregressive approach.
\newblock In {\em Proceedings of the IEEE conference on computer vision and
  pattern recognition}, pages 1370--1377, 2014.

\bibitem{zhu2017unpaired}
Jun-Yan Zhu, Taesung Park, Phillip Isola, and Alexei~A Efros.
\newblock Unpaired image-to-image translation using cycle-consistent
  adversarial networks.
\newblock In {\em Proceedings of the IEEE international conference on computer
  vision}, pages 2223--2232, 2017.

\bibitem{zimmermann2017learning}
Christian Zimmermann and Thomas Brox.
\newblock Learning to estimate 3d hand pose from single rgb images.
\newblock In {\em Proceedings of the IEEE international conference on computer
  vision}, pages 4903--4911, 2017.

\end{thebibliography}

\end{document}